\pdfoutput=1
\documentclass{article}
\usepackage[numbers,sort&compress]{natbib}
\usepackage{booktabs}
\usepackage{PRIMEarxiv}
\usepackage[table,xcdraw]{xcolor}
\usepackage[utf8]{inputenc}
\usepackage[T1]{fontenc}
\usepackage{url}
\usepackage{amsfonts}
\usepackage{nicefrac}
\usepackage{microtype}
\usepackage{fancyhdr}
\usepackage{graphicx}
\usepackage{adjustbox}
\usepackage{amsmath}
\usepackage{amssymb}
\usepackage{multirow}
\usepackage{float}
\usepackage{xspace}
\usepackage{caption}
\usepackage[shortlabels,inline]{enumitem}

\usepackage{currfile}   %
\usepackage{multirow}
\usepackage{amsthm}
\usepackage{pifont}     %
\usepackage{needspace}  %
\usepackage{tikz}
\usetikzlibrary{positioning,arrows.meta,fit,backgrounds,calc,shapes.misc}

\newcommand{\ci}[1]{{\scriptsize[#1]}}

\newcommand{\ours}{Mirror-JEPA\xspace}
\newcommand{\Native}{Native\xspace}
\newcommand{\Change}{Change\xspace}
\newcommand{\ChangeKF}{Change-KF\xspace}        %
\newcommand{\ChangeM}{Change-M\xspace}          %
\newcommand{\Conf}{Confidence\xspace}
\newcommand{\EgoTemp}{Ego-Temporal\xspace}
\newcommand{\Rand}{Random\xspace}
\newcommand{\cooponly}{\textsc{Coop-only}\xspace}
\newcommand{\moving}{\textsc{Moving}\xspace}

\newcommand{\apstrict}{3D AP@0.7\xspace}

\newcommand{\Hold}{\textsc{hold}\xspace}
\newcommand{\Pred}{\textsc{pred}\xspace}
\newcommand{\Clear}{\textsc{clear}\xspace}
\newcommand{\Send}{\textsc{send}\xspace}

\newcommand{\bz}{\mathbf{z}}
\newcommand{\bh}{\mathbf{h}}
\newcommand{\bw}{\mathbf{w}}
\newcommand{\M}{\mathcal{M}}
\newcommand{\hM}{\widehat{\mathcal{M}}}

\theoremstyle{plain}
\newtheorem{proposition}{Proposition}

\theoremstyle{remark}

\usepackage{hyperref}
\hypersetup{
  pdftitle={Fresh, Not Full: Closed-Loop Latent Prediction for Cooperative Perception over Real C-V2X Links},
  pdfauthor={Yunxiang Yang, Ningning Xu, Jidong J. Yang}
}
\usepackage[capitalize]{cleveref}
\crefname{section}{Sec.}{Secs.}
\Crefname{section}{Section}{Sections}
\Crefname{table}{Table}{Tables}
\crefname{table}{Tab.}{Tabs.}

\title{Fresh, Not Full: Closed-Loop Latent Prediction for Cooperative Perception over Real C-V2X Links}

\author{
 Yunxiang Yang, Ningning Xu, Jidong J. Yang$^{*}$ \\
Smart Mobility and Infrastructure Lab\\
College of Engineering, University of Georgia \\
  \texttt{\{yyang117, Ningning.Xu, Jidong.Yang\}@uga.edu} \\
}

\begin{document}
\maketitle

\renewcommand*{\thefootnote}{\fnsymbol{footnote}}
\footnotetext{$^{*}$Corresponding author.}

\begin{abstract}
Cooperative perception enables vehicles to share intermediate features, but over real C-V2X links a full feature message can be nearly one second old by the time it is received and fused. On the CooperScene benchmark, this delay reduces the CoSDH detector's \apstrict by 6.8 points. Selecting what to transmit by detection confidence, a common criterion in communication-efficient methods, recovers less than 5\% of this loss. In contrast, transmitting only the feature cells that differ from the receiver's current state recovers approximately 80\% of the loss while using just 6\% of the bytes. Realizing this advantage requires the sender to know what the receiver currently holds. We address this challenge by maintaining an exact mirror of the receiver's feature state at each sender. For every cell, the sender determines whether the receiver should  retain its existing feature, clear it, replace it with a latent prediction generated jointly by both ends, or receive an updated one. Crucially, a prediction is used only when the sender has verified its consistency with the current feature, at a cost of approximately one byte per cell compared with twelve bytes for a full feature. Under an equal-byte keyframe setting, the resulting protocol, \ours, outperforms a causal keyframe baseline on the held-out test set, improving BEV  AP@0.7 by 2.2 points and 3D AP@0.5 by 4.0 points, while satisfying all criteria established before test evaluation. Compared with transmitting full feature messages every frame, \ours reduces communication by a factor of 4.5 and achieves higher scores on three of four test metrics. It also reduces false positives by 89\% at locations previously occupied by vehicles.
\end{abstract}


\section{Introduction}
\label{sec:intro}
\begin{figure}[!htbp]
\centering
\includegraphics[width=\linewidth]{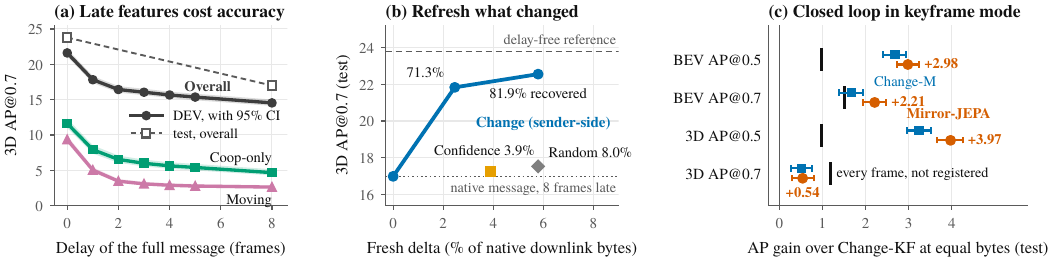}
\caption{\textbf{Fresh, not full.} (a) DEV set: \apstrict of CoSDH as a function of the delay of its full messages (95\% bootstrap CIs); open squares: the test set without delay and at the eight-frame delay that CooperScene reports for C-V2X. \moving: objects displaced by at least 0.5\,m over 0.5\,s; \cooponly: objects that only collaborators observe (Supp.~\ref{sec:supp_diag}). (b) Test set: a small on-time delta added to the delayed messages, with cells selected by sender-side change, by detection confidence, or at random; labels: share of the delay loss recovered. (c) Test set, keyframe mode at equal bytes: improvements of the closed loop (\ChangeM) and of its latent predictor (\ours) over the causal keyframe-change baseline (\ChangeKF), with 95\% CIs over queries (robustness: \cref{sec:exp_ladder}). Black marks: the every-frame stream.}
\label{fig:teaser}
\end{figure}

A vehicle can perceive only the portion of its surroundings covered by its onboard sensors. Cooperative perception extends this limited view by allowing vehicles and roadside units (RSUs) to share intermediate bird's-eye-view (BEV) features~\cite{wang2020v2vnet,xu2022v2xvit,hu2022where2comm}. Existing methods are typically evaluated under fixed bandwidth budgets or assumed communication latencies, which may not reflect real-world network conditions. The recently introduced CooperScene benchmark~\cite{wu2026cooperscene} addresses this limitation by measuring communication over a real C-V2X link among three connected vehicles and one RSU, providing model-specific throughput and message latency measurements. To our knowledge, no other dataset records C-V2X throughput. For CoSDH~\cite{xu2025cosdh}, the messages that a vehicle receives from its three collaborators total 178.5\,KB and take 893\,ms to arrive. At 10\,Hz, the receiver fuses features that are eight frames old.

This delay is costly. On the test set, it lowers the \apstrict of CoSDH from 23.78 to 16.99 (\cref{fig:teaser}a). On the validation set, one frame of delay already costs about half of the loss at eight frames (\cref{fig:teaser}a). We therefore ask which content is essential to arrive on time. We keep the delayed messages as a background and add a small stream that arrives within the frame it describes, and we compare criteria for choosing its cells (\cref{sec:analysis}). Detection confidence, the spatial confidence map of Where2comm~\cite{hu2022where2comm}, recovers less than 5\% of the lost AP, while random cells recover 8\%. Sending the cells that changed the most relative to what the receiver already holds recovers about 80\% with 6\% of the bytes (\cref{fig:teaser}b).

Sending only changes requires the sender to know the receiver’s current state. As small updates accumulate at the receiver, the sender must continuously track them. We address this by having each sender maintain a mirror of the receiver’s state. Because the receiver’s update rule is deterministic given the messages and their arrival frames, and the sender knows both, the sender can maintain an exact replica of the receiver’s state. Across all 95,446 link-frames in our test replay, the mirror remained perfectly synchronized with the actual receiver state.

With the mirror, the sender can decide, for every cell, what action the receiver should take (\cref{fig:method}). The receiver can retain its current feature, clear it, replace it with the output of a latent predictor executed at both ends, or receive a new feature. The predictor is a lightweight JEPA-style network that transports held features along a learned displacement field. 
Its output is used only when the sender verifies it against its current feature, much like an encoder in a predictive coder runs the decoder's predictor ~\cite{jayant1984digital,wiegand2003h264}. This verification is critical: applying the same type of prediction without verification reduces AP by 6.7 points on a diagnostic subset. A verified prediction requires only about one byte of communication, compared with twelve bytes for transmitting a feature.

The same communication loop can also accommodate links that cannot support full-message transmission at every frame.  In keyframe mode, each collaborator sends a full message every ten frames and transmits only verified increments in between. We refer to this protocol as \ours (\cref{fig:teaser}c). Our contributions are summarized as follows.
\begin{itemize}[leftmargin=*,itemsep=1pt,topsep=2pt]
\item We show that under the C-V2X latency measured by CooperScene, transmitting changes relative to the receiver's current state is substantially more effective than selecting features by detection confidence. Using only 6\% of the native message size, change-ranked cells recover 78.5--81.9\% of the AP lost to an eight-frame delay across four metrics, whereas confidence-ranked and randomly selected cells recover at most 8.1\% (\cref{tab:main}).
\item We introduce a sender-side mirror that exactly reproduces the receiver's state, with zero discrepancy across 95,446 link-frames in the test replay.  This mirror enables the sender to select among four per cell actions and verify predictions before they are applied by the receiver, at a communication cost of approximately one byte per cell (\cref{sec:method}).
\item \ours, the keyframe protocol built on this loop, outperforms a causal keyframe baseline at equal bytes on the test set (+3.0 BEV AP@0.5, +2.2 BEV AP@0.7, +4.0 3D AP@0.5 and +0.5 \apstrict) and passes every criterion specified before test evaluation. The closed loop accounts for 75--94\% of the improvement, while the predictor contributes  up to an additional 0.73 AP at no extra communication cost (\cref{tab:ladder}).
\item We trace these gains to feature freshness. Improvements are concentrated on moving objects, and false positives at vacated positions decrease by about 90\% relative to both the keyframe baseline and the every-frame stream.  Compared with transmitting full CoSDH messages over the same link, the protocol improves BEV AP@0.5 by 14.5 points and \apstrict by 4.9 points while using approximately one-seventh of the downlink bandwidth. Compared with streaming full messages every frame, it reduces communication by a factor of 4.5, achieves higher scores on three of four test metrics, and incurs at most a 0.8-point AP reduction on the remaining metric (\cref{sec:exp_gain,sec:exp_bytes}).
\end{itemize}

\def\zxa{0.3398}\def\zxb{0.5848}\def\zya{0.4012}\def\zyb{0.7122}
\definecolor{mjfeat}{HTML}{3A6B9A}   %
\definecolor{mjpred}{HTML}{E69F00}   %
\definecolor{mjsend}{HTML}{D55E00}   %
\definecolor{mjclear}{HTML}{56B4E9}  %
\definecolor{mjhold}{HTML}{E4E4E4}   %
\definecolor{mjsl}{HTML}{EEF3F9}     %
\definecolor{mjrl}{HTML}{FBF3E8}     %
\definecolor{mjmute}{HTML}{5F5F5F}
\definecolor{scgt}{HTML}{0E9F8E}     %
\definecolor{scpred}{HTML}{D62728}   %
\definecolor{scptr}{HTML}{8C8C8C}    %
\definecolor{scptc}{HTML}{9CC3EA}    %
\definecolor{sctri}{HTML}{1F4E8C}    %
\def\cs{0.25}%
\newcommand{\stepn}[1]{{\fontsize{8.5}{8.5}\selectfont\ding{#1}}}%
\newcommand{\bev}[3]{%
  \fill[white] (#2,#3) rectangle ++(7*\cs,4*\cs);
  \draw[black!13, line width=0.25pt, xshift=#2 cm, yshift=#3 cm, step=\cs] (0,0) grid (7*\cs,4*\cs);
  \draw[black!60, line width=0.5pt] (#2,#3) rectangle ++(7*\cs,4*\cs);
  \coordinate (#1-sw) at (#2,#3); \coordinate (#1-nw) at (#2,{#3+4*\cs});
  \coordinate (#1-se) at ({#2+7*\cs},#3); \coordinate (#1-ne) at ({#2+7*\cs},{#3+4*\cs});
  \coordinate (#1-w) at (#2,{#3+2*\cs}); \coordinate (#1-e) at ({#2+7*\cs},{#3+2*\cs});
  \coordinate (#1-n) at ({#2+3.5*\cs},{#3+4*\cs}); \coordinate (#1-s) at ({#2+3.5*\cs},#3);
}
\newcommand{\cell}[4][mjfeat]{\fill[#1] ($(#2-sw)+({(#3-1)*\cs+0.018},{(#4-1)*\cs+0.018})$) rectangle ++(\cs-0.036,\cs-0.036);}
\newcommand{\cellL}[5][mjfeat]{\cell[#1]{#2}{#3}{#4}\node[font=\scriptsize\bfseries, text=white, inner sep=0pt] at ($(#2-sw)+({(#3-0.5)*\cs},{(#4-0.5)*\cs})$) {#5};}
\newcommand{\cellO}[4][mjclear]{\draw[#1, line width=0.55pt, densely dashed] ($(#2-sw)+({(#3-1)*\cs+0.035},{(#4-1)*\cs+0.035})$) rectangle ++(\cs-0.07,\cs-0.07);}
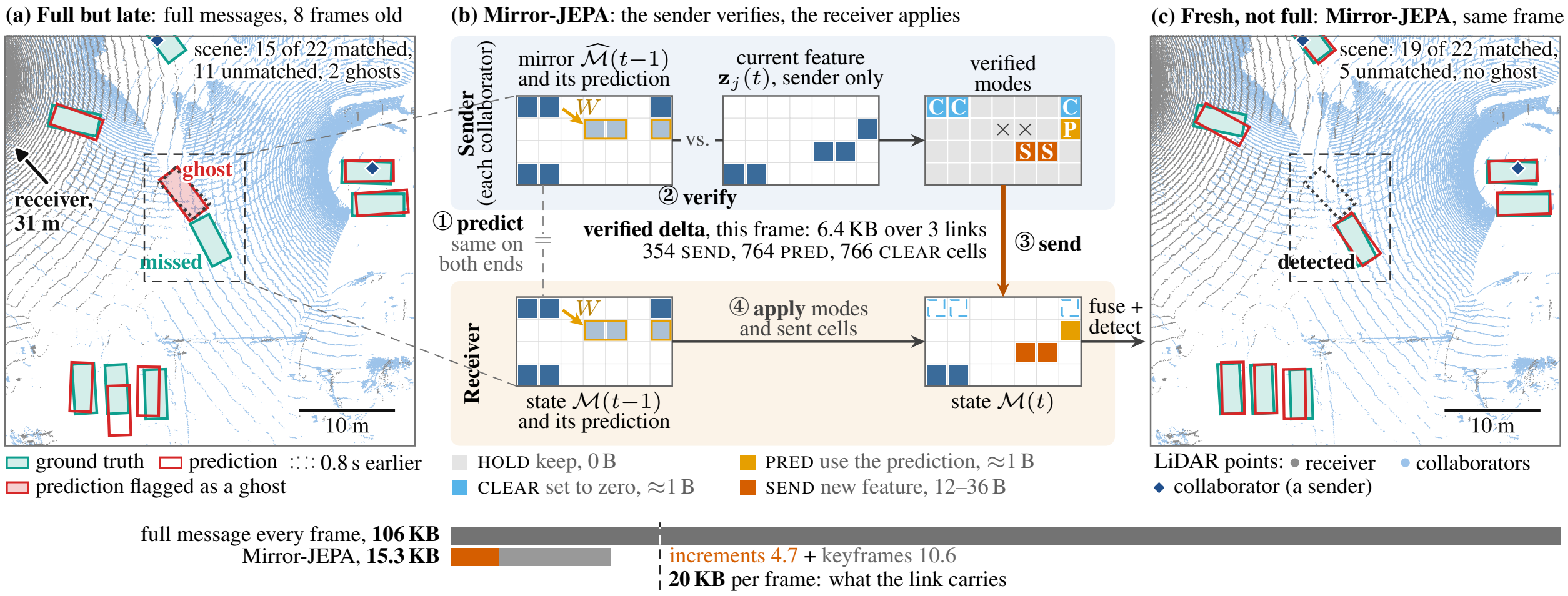
\begin{figure*}[!htbp]
\centering
\resizebox{\textwidth}{!}{%
\begin{tikzpicture}[x=1cm,y=1cm, font=\scriptsize, >={Stealth[length=1.6mm,width=1.4mm]},
  lbl/.style={font=\scriptsize, inner sep=1.5pt, align=center},
  mut/.style={font=\scriptsize, text=mjmute, inner sep=1.5pt, align=center},
  leg/.style={font=\scriptsize, inner sep=1pt, anchor=west},
  flow/.style={->, line width=0.7pt, black!75}]
\def\pw{4.6}%
\node[anchor=south west, inner sep=0] at (0,1.05) {\includegraphics[width=\pw cm]{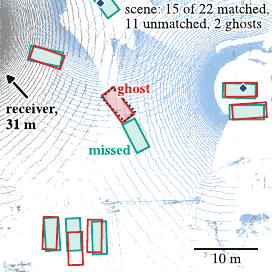}};
\node[anchor=south west, inner sep=0] at (12.86,1.05) {\includegraphics[width=\pw cm]{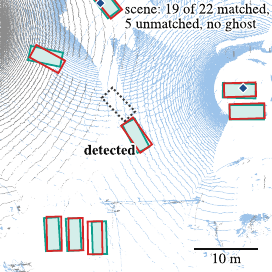}};
\foreach \x in {0,12.86}{
  \draw[black!55, line width=0.5pt] (\x,1.05) rectangle ++(\pw,\pw);
  \draw[black!80, line width=0.5pt, densely dashed] ({\x+\zxa*\pw},{1.05+\zya*\pw}) rectangle ({\x+\zxb*\pw},{1.05+\zyb*\pw});
}
\node[lbl, anchor=south west, inner sep=0pt] at (0,5.75) {\textbf{(a) Full but late}: full messages, 8 frames old};
\node[lbl, anchor=south west, inner sep=0pt] at (12.86,5.75) {\textbf{(c) Fresh, not full}: \textbf{\ours}, same frame};
\begin{scope}[shift={(0,0.86)}]
  \draw[scgt, fill=scgt!18, line width=0.6pt] (0.02,-0.06) rectangle ++(0.24,0.12); \node[leg] at (0.30,0) {ground truth};
  \draw[scpred, line width=0.6pt] (1.74,-0.06) rectangle ++(0.24,0.12); \node[leg] at (2.02,0) {prediction};
  \draw[black!75, line width=0.6pt, dotted] (3.22,-0.06) rectangle ++(0.24,0.12); \node[leg] at (3.50,0) {0.8\,s earlier};
  \draw[scpred, fill=scpred!22, line width=0.6pt] (0.02,-0.34) rectangle ++(0.24,0.12); \node[leg] at (0.30,-0.28) {prediction flagged as a ghost};
\end{scope}
\begin{scope}[shift={(12.86,0.86)}]
  \node[leg] at (0,0) {LiDAR points:};
  \fill[scptr] (1.62,0) circle (0.05); \node[leg] at (1.70,0) {receiver};
  \fill[scptc] (2.86,0) circle (0.05); \node[leg] at (2.94,0) {collaborators};
  \fill[sctri, draw=white, line width=0.3pt] (0.095,-0.35) -- ++(0.07,0.07) -- ++(-0.07,0.07) -- ++(-0.07,-0.07) -- cycle; \node[leg] at (0.22,-0.28) {collaborator (a sender)};
\end{scope}
\fill[mjsl, rounded corners=3pt] (5.00,3.70) rectangle (12.46,5.65);
\fill[mjrl, rounded corners=3pt] (5.00,1.05) rectangle (12.46,2.90);
\node[rotate=90, lbl] at (5.30,4.675) {\textbf{Sender}\\[-2pt]{(each collaborator)}};
\node[rotate=90, lbl] at (5.24,1.975) {\textbf{Receiver}};
\bev{sA}{5.74}{3.98} \bev{sZ}{8.06}{3.98} \bev{sQ}{10.33}{3.98}
\bev{rA}{5.74}{1.72} \bev{rN}{10.33}{1.72}
\foreach \g in {sA,rA}{
  \cell{\g}{1}{1} \cell{\g}{2}{1} \cell{\g}{1}{4} \cell{\g}{2}{4} \cell{\g}{7}{4}
  \cell[mjfeat!42]{\g}{4}{3} \cell[mjfeat!42]{\g}{5}{3} \cell[mjfeat!42]{\g}{7}{3}
  \draw[mjpred, line width=0.6pt] ($(\g-sw)+(3*\cs+0.02,2*\cs+0.02)$) rectangle ++(2*\cs-0.04,\cs-0.04);
  \draw[mjpred, line width=0.6pt] ($(\g-sw)+(6*\cs+0.02,2*\cs+0.02)$) rectangle ++(\cs-0.04,\cs-0.04);
  \draw[->, mjpred, line width=1.0pt] ($(\g-sw)+(2.08*\cs,3.42*\cs)$) -- ($(\g-sw)+(3.02*\cs,2.72*\cs)$);
  \node[font=\scriptsize, text=mjpred!80!black, inner sep=0pt] at ($(\g-sw)+(3.25*\cs,3.5*\cs)$) {$W$};
}
\cell{sZ}{1}{1} \cell{sZ}{2}{1} \cell{sZ}{5}{2} \cell{sZ}{6}{2} \cell{sZ}{7}{3}
\fill[mjhold] ($(sQ-sw)+(0.01,0.01)$) rectangle ++(7*\cs-0.02,4*\cs-0.02);
\draw[white, line width=0.45pt, xshift=10.33cm, yshift=3.98cm, step=\cs] (0,0) grid (7*\cs,4*\cs);
\draw[black!60, line width=0.5pt] (sQ-sw) rectangle ++(7*\cs,4*\cs);
\cellL[mjclear]{sQ}{1}{4}{C} \cellL[mjclear]{sQ}{2}{4}{C} \cellL[mjclear]{sQ}{7}{4}{C} \cellL[mjsend]{sQ}{5}{2}{S} \cellL[mjsend]{sQ}{6}{2}{S} \cellL[mjpred]{sQ}{7}{3}{P}
\node[font=\scriptsize, text=black!80, inner sep=0pt] at ($(sQ-sw)+(3.5*\cs,2.5*\cs)$) {$\times$};
\node[font=\scriptsize, text=black!80, inner sep=0pt] at ($(sQ-sw)+(4.5*\cs,2.5*\cs)$) {$\times$};
\cell{rN}{1}{1} \cell{rN}{2}{1} \cellO{rN}{1}{4} \cellO{rN}{2}{4} \cellO{rN}{7}{4} \cell[mjsend]{rN}{5}{2} \cell[mjsend]{rN}{6}{2} \cell[mjpred]{rN}{7}{3}
\node[lbl, anchor=south] at ($(sA-n)+(0,0.0)$) {mirror $\hM(t{-}1)$\\[-1.5pt]and its prediction};
\node[lbl, anchor=south] at ($(sZ-n)+(0,0.0)$) {current feature\\[-1.5pt]$\bz_j(t)$, sender only};
\node[lbl, anchor=south] at ($(sQ-n)+(0,0.0)$) {verified\\[-1.5pt]modes};
\node[lbl, anchor=north] at ($(rA-s)+(0,0.0)$) {state $\M(t{-}1)$\\[-1.5pt]and its prediction};
\node[lbl, anchor=north] at ($(rN-s)+(0,0.0)$) {state $\M(t)$};
\draw[line width=0.7pt, black!75] (sA-e) -- (sZ-w) node[midway, fill=mjsl, inner sep=1.5pt, font=\scriptsize] {vs.};
\draw[flow] (sZ-e) -- (sQ-w);
\node[lbl, anchor=north, inner sep=0pt] at ($(sA-e)!0.5!(sZ-w)+(0,-0.50)$) {\stepn{173}\,\textbf{verify}};
\draw[flow] (rA-e) -- node[lbl, above, pos=0.5] {\stepn{175}\,\textbf{apply} modes\\[-1.5pt]and sent cells} (rN-w);
\draw[black!45, line width=0.5pt, densely dashed] ($(sA-sw)+(0.30,0)$) -- ($(rA-nw)+(0.30,0)$) node[pos=0.50, fill=white, inner sep=1pt, font=\scriptsize] {$=$};
\node[lbl, anchor=east, align=right] at ($(sA-sw)!0.50!(rA-nw)+(0.14,0)$) {\stepn{172}\,\textbf{predict}\\[-1.5pt]{\color{mjmute}same on}\\[-1.5pt]{\color{mjmute}both ends}};
\draw[->, line width=1.5pt, mjsend!85!black] (sQ-s) -- (rN-n);
\node[lbl, anchor=east, align=right] at ($(sQ-s)!0.50!(rN-n)+(-0.12,0)$) {\textbf{verified delta}, this frame: 6.4\,KB over 3 links\\[-1pt]354 \Send, 764 \Pred, 766 \Clear cells};
\node[lbl, anchor=west, inner sep=0pt] at ($(sQ-s)!0.50!(rN-n)+(0.12,0)$) {\stepn{174}\,\textbf{send}};
\draw[black!55, line width=0.4pt, densely dashed] ({\zxb*\pw},{1.05+\zyb*\pw}) -- (sA-nw);
\draw[black!55, line width=0.4pt, densely dashed] ({\zxb*\pw},{1.05+\zya*\pw}) -- (rA-sw);
\draw[flow] (rN-e) -- (12.80,{1.72+2*\cs});
\node[lbl, anchor=south west, inner sep=1.5pt, align=left] at ($(rN-e)+(0.03,0.03)$) {fuse +\\[-1.5pt]detect};
\node[lbl, anchor=south west, inner sep=0pt] at (5.00,5.75) {\textbf{(b) \ours}: the sender verifies, the receiver applies};
\foreach \x/\y/\c/\n/\d in {5.02/0.86/mjhold/\Hold/{keep, 0\,B}, 8.25/0.86/mjpred/\Pred/{use the prediction, $\approx$1\,B},
                            5.02/0.58/mjclear/\Clear/{set to zero, $\approx$1\,B}, 8.25/0.58/mjsend/\Send/{new feature, 12--36\,B}}{
  \fill[\c] (\x,\y-0.085) rectangle ++(0.17,0.17);
  \node[lbl, anchor=west, inner sep=1pt] at (\x+0.24,\y) {\n\ {\color{mjmute}\d}};
}
\def\bx{5.00}\def\bw{12.46}
\begin{scope}[yshift=-0.30cm]
\fill[black!55] (\bx,0.24) rectangle ++(\bw,0.20);
\fill[mjsend] (\bx,0.00) rectangle ++({4.7/106*\bw},0.20);
\fill[black!40] ({\bx+4.7/106*\bw},0.00) rectangle ++({10.6/106*\bw},0.20);
\draw[black!85, line width=0.6pt, densely dashed] ({\bx+20/106*\bw},-0.27) -- ({\bx+20/106*\bw},0.24);
\draw[white, line width=0.6pt, densely dashed] ({\bx+20/106*\bw},0.24) -- ({\bx+20/106*\bw},0.44);
\draw[black!85, line width=0.6pt, densely dashed] ({\bx+20/106*\bw},0.44) -- ({\bx+20/106*\bw},0.49);
\node[lbl, anchor=east, inner sep=2pt] at (\bx-0.04,0.34) {full message every frame, \textbf{106\,KB}};
\node[lbl, anchor=east, inner sep=2pt] at (\bx-0.04,0.10) {\ours, \textbf{15.3\,KB}};
\node[lbl, anchor=west, inner sep=2pt] at ({\bx+20/106*\bw+0.03},0.10) {{\color{mjsend}increments 4.7} + {\color{black!60}keyframes 10.6}};
\node[lbl, anchor=west, inner sep=2pt] at ({\bx+20/106*\bw+0.02},-0.17) {\textbf{20\,KB} per frame: what the link carries};
\end{scope}
\end{tikzpicture}}
\caption{\textbf{\ours on one test query} (target selected by the rule of Supp.~\ref{sec:supp_qual}). Points represent current LiDAR returns in (a) and (c) : gray for the receiver (located outside the view, indicated by the arrow) and blue for the collaborators. \textbf{(a)} Full messages fused eight frames late: a vehicle that moved 5.6\,m in 0.8\,s is missed (3D IoU 0.00), and a ghost remains at its previous position (shaded in red). \textbf{(b)} Schematic of the communication loop on one link (the cells of this frame are shown in Supp.~Fig.~\ref{fig:supp_cells}); each collaborator acts as a sender. (1) Both ends execute the predictor $W$ on the same state, which the sender maintains as a mirror. (2) For each cell, the sender verifies whether holding, prediction, or clearing reproduces its current feature. A prediction that fails verification (denoted by $\times$) is never applied; here, prediction fails on the moving vehicle, so its cells are transmitted. (3) The sender transmits the selected modes and the current features of cells that cannot be reproduced by any mode. (4) The receiver applies the received updates. \textbf{(c)} The same frame reconstructed with \ours: the vehicle is detected (IoU 0.74), with no remaining ghost. Across the test set, among 14,144 objects moving at 2--10\,m/s, 1,982 improve from an IoU below 0.5 in (a) to at least 0.7 in (c), whereas 42 exhibit the opposite change. Bars at the bottom show test-set averages.}
\label{fig:method}
\end{figure*}

\section{Related Work}
\label{sec:related}

\paragraph{Communication-efficient cooperative perception.}
Intermediate fusion shares BEV features among agents and fuses them at the receiver~\cite{wang2020v2vnet,xu2022v2xvit,xu2022opv2v}. To satisfy a bandwidth budget, Where2comm~\cite{hu2022where2comm} transmits only the spatially confident cells requested by the receiver, and CoSDH~\cite{xu2025cosdh} matches the sender's supply to the receiver's demand. CodeFilling~\cite{hu2024codefilling} replaces features with codebook indices, and V2X-DSC~\cite{zeng2026v2xdsc} treats the receiver's own features as decoder side information~\cite{wyner1976rate}. CooperTrim~\cite{mukhopadhyay2026coopertrim} allows the receiver to request regions in which its own features change over time. These methods select the transmitted content according to spatial importance or receiver-side uncertainty, and their evaluations treat message size and latency as separate quantities. V2X-ReaLO~\cite{xiang2025v2xrealo} measures both online over a Wi-Fi link. On a real link, however, message size determines latency, and these criteria recover far less of the delay-induced loss than sender-side change (\cref{sec:analysis}). CoST~\cite{tang2025cost} is the closest in spirit, as it transmits dynamic content relative to earlier frames. In contrast, we measure change against the receiver's actual state, which the sender reconstructs, and we adopt a CoST-style keyframe protocol as a baseline.

\paragraph{Latency compensation.}
Several methods accept that messages arrive late and correct the stale features at the receiver. SyncNet~\cite{lei2022syncnet} and How2comm~\cite{yang2023how2comm} learn temporal compensation modules, and FFNet~\cite{yu2023ffnet} transmits a feature flow that the receiver extrapolates. CoBEVFlow~\cite{wei2023cobevflow} moves features along an estimated BEV flow, and TraF-Align~\cite{song2025trafalign} aligns them along predicted trajectories. All of these methods predict at the receiver without feedback on the prediction error. In \ours, the receiver applies a prediction only where the sender has verified it.

\paragraph{Freshness and closed-loop prediction.}
Age of information quantifies the staleness of a receiver's view~\cite{kaul2012realtime,sun2017update}, and Ma et al.~\cite{ma2026unseen} schedule updates for regions that the receiver has not observed. In differential pulse-code modulation (DPCM) and video coding, the encoder runs the decoder's predictor and transmits the residual~\cite{jayant1984digital,wiegand2003h264}. Concurrently, WorldGT~\cite{gong2026guess} maintains a twin of a single video receiver at the sender and verifies the predictions of a shared world model. We study this closed loop for multi-agent LiDAR perception, in which the link converts message size into whole frames of delay.

\paragraph{World models and JEPA.}
Joint-embedding predictive architectures (JEPA) predict representations rather than pixels~\cite{lecun2022path,assran2023ijepa,bardes2024vjepa}. Like V-JEPA 2-AC~\cite{assran2025vjepa2} and DINO-WM~\cite{zhou2024dinowm}, our predictor $W$ is trained on top of a frozen encoder. JEPA and reconstruction objectives have also been used to pretrain LiDAR and V2X encoders~\cite{zhu2026adljepa,zhu2026adlistjepa,zhao2024coopre,mayumu2026v2xjepa}, and World4V2X~\cite{wang2026world4v2x} incorporates a world model at inference time. In contrast, our predictor operates at both ends of the link within a verification loop.

\section{Cooperative Perception on a Real C-V2X Link}
\label{sec:analysis}

\subsection{Setting}
\label{sec:setting}

\paragraph{Data and link model.}
CooperScene~\cite{wu2026cooperscene} records three connected vehicles (CAVs) and one RSU at 10\,Hz across 24 scenes. Each agent is equipped with LiDAR and a commercial C-V2X radio. CooperScene reports that the CoSDH~\cite{xu2025cosdh} messages received by a vehicle from its three collaborators total 178.5\,KB (59.5\,KB per collaborator) and require 893\,ms to arrive, which corresponds to a throughput of 1.6\,Mbps (1\,KB $=$ 1,000\,B).  Thus, messages are approximately eight frames old when they are fused. We adopt this communication model. Among the six detectors evaluated in the benchmark, CoSDH produces the smallest messages and achieves the highest accuracy over C-V2X with three collaborators~\cite{wu2026cooperscene} (\cref{tab:benchmark}); we therefore build on it. We refer to the full CoSDH message for a given frame as the \emph{native} message. Native messages are used eight frames after capture.  For a stream transmitting a fraction $r$ of the native message size, we model the delay as $\lfloor 8.925\,r\rfloor$ frames. Consequently, messages are available in the frame they describe when $r<11.2\%$. This delay is fixed for each condition based on its mean message size, rather than computed separately for each message. Receiver requests travel in the reverse direction and are accounted for separately (Supp.~\ref{sec:supp_protocol}).  Communication costs are measured per transmitted cell and include the compressed latent and a 4-byte index.  Each message is computed from the sender's point cloud at capture time and expressed in the receiver's coordinate frame at the time of fusion. We therefore assume that both ends know the current relative pose at every frame; the receiver's request carries its pose.

\begin{table}[!htbp]
\caption{\textbf{Cooperative detection over C-V2X.} Top: the benchmark's results for three collaborators under its own protocol~\cite{wu2026cooperscene}. Bottom: our replay of the released CoSDH checkpoint on the test set (3,936 queries, 2.5 collaborators on average; \cref{sec:exp}). The two blocks differ in test set and metric and are not comparable with each other; the CoSDH rows show the offset. Size: data shared with a receiver per frame. Latency: of the full messages or keyframes and, after the slash, of the on-time stream. Under \apstrict the order of the \Change delta and \ours reverses (\cref{tab:ladder}). $^\dagger$Not achievable on the measured link.}
\label{tab:benchmark}
\centering
\footnotesize
\setlength{\tabcolsep}{1.8pt}
\begin{adjustbox}{max width=\linewidth}
\begin{tabular}{@{}lrrcc@{}}
\toprule
& Size & Latency & & \\
Method & (MB) & (ms) & unlimited & C-V2X \\
\midrule
\multicolumn{5}{@{}l}{\textit{Reported by the benchmark~\cite{wu2026cooperscene}, three collaborators: mAP@0.5 / 0.7}} \\
V2VNet~\cite{wang2020v2vnet} & 30.0 & 107,432 & 0.75 / 0.50 & 0.23 / 0.13 \\
V2X-ViT~\cite{xu2022v2xvit} & 1.014 & 9,976 & 0.74 / 0.50 & 0.52 / 0.33 \\
V2VAM~\cite{li2023v2vam} & 1.269 & 11,302 & 0.78 / 0.53 & 0.54 / 0.34 \\
CoBEVT~\cite{xu2022cobevt} & 1.500 & 12,381 & 0.75 / 0.54 & 0.50 / 0.32 \\
ERMVP~\cite{zhang2024ermvp} & 1.038 & 5,190 & \textbf{0.81 / 0.57} & 0.56 / 0.36 \\
CoSDH~\cite{xu2025cosdh} & \textbf{0.1785} & \textbf{893} & 0.73 / 0.52 & \textbf{0.61 / 0.41} \\
\midrule
\multicolumn{5}{@{}l}{\textit{Our replay of the CoSDH checkpoint, test set: BEV AP@0.5 / 0.7}} \\
CoSDH, native message & 0.106 & 893 & 70.27 / 46.96$^\dagger$ & 54.43 / 34.20 \\
+ \Change delta (ours) & 0.112 & 893 / $<$100 & -- & 66.87 / 44.58 \\
Keyframes, CoST-style~\cite{tang2025cost} & 0.0153 & 893 / $<$100 & -- & 65.90 / 43.07 \\
Keyframes, \ours (ours) & 0.0153 & 893 / $<$100 & -- & \textbf{68.88 / 45.29} \\
\bottomrule
\end{tabular}
\end{adjustbox}
\end{table}

\paragraph{Detector, queries and metrics.}
We use the CoSDH checkpoint released with the benchmark (PointPillar encoder, 0.4\,m voxels) without fine-tuning, with a score threshold of 0.20 and an NMS threshold of 0.15. Queries consist of frames from the three CAVs (1,312 frames per vehicle on the test set); the RSU acts only as a sender. Each query has up to three collaborators, with an average of 2.5. We report AP at IoU 0.5 and 0.7 for both BEV IoU and 3D IoU, where 3D IoU accounts for both the BEV polygon intersection and the vertical overlap. \apstrict is the preregistered primary metric of the keyframe comparisons. We additionally report AP across four speed bins. Confidence intervals are 95\% paired bootstrap intervals computed over queries using 1,000 resamples.

\paragraph{Data splits.}
The public release used in our study contains 14 scenes organized into three folders: ten training scenes, two validation scenes, and two testing scenes. The test set (scenes 2 and 8; 3,936 queries) is used to report the primary results, with each registered configuration evaluated on it once. The DEV set (the validation scenes 13 and 14; 2,406 queries) is used for development and confirmation. A \emph{diagnostic subset} of 128 queries drawn from three training scenes (10 to 12) is used exclusively for controlled comparisons (Supp.~\ref{sec:supp_diag}). The predictor is trained on the other seven training scenes.

\paragraph{Every-frame protocol.}
In the protocol of \cref{tab:main}, each collaborator streams its native message, which the receiver processes with an eight-frame delay. A \emph{delta} stream supplements this delayed message with the current values of a selected subset of cells, applied in the frame they represent. All conditions share the same delayed background and differ only in which cells are refreshed (Supp.~\ref{sec:supp_protocol}).

\subsection{What to refresh}
\label{sec:select}

On the test set, the eight-frame delay reduces performance by  15.8 BEV AP@0.5, 12.8 BEV AP@0.7, 13.8 3D AP@0.5 and 6.8 \apstrict (\cref{tab:main}; Supp.~\ref{sec:supp_everyframe}). On DEV, one frame of delay already costs 3.8 of the 7.1 points of \apstrict lost at eight frames (\cref{fig:teaser}a; Supp.~\ref{sec:supp_protocol}).

Let $\bz_k(t)$ denote the sender's current feature in BEV cell $k$ and let $\bh_k$ denote the feature currently held by the receiver for that cell, which comes from a message transmitted eight frames earlier. The delta stream refreshes cells whose scores exceed a threshold. Each threshold is defined by a quantile $\tau$ of the score distribution over requested and observed cells, calibrated on the diagnostic subset, and then held fixed. We consider four scoring criteria: two computed at the sender and two at the receiver.  \Change measures the difference between the sender's current feature and the receiver's stored feature, computed as $c_k=\frac{1}{C}\lVert\bz_k(t)-\bh_k\rVert_1$. \Conf uses the sender's spatial confidence map, computed as in Where2comm~\cite{hu2022where2comm}: the maximum class score at each cell before fusion, smoothed with a Gaussian filter. CoSDH uses the same map to determine which features a sender transmits. The two receiver-side criteria are: \EgoTemp, which measures the change in the receiver's own feature between consecutive frames (following a CooperTrim-style criterion~\cite{mukhopadhyay2026coopertrim}), and JEPA error, which estimates the expected error from retaining the current feature, $\lVert P_k-\bh_k\rVert^2+\sigma_k^2$, using the predictor described in  \cref{sec:predictor}. Finally, \Rand randomly selects the same number of cells as \Change.

\Cref{sec:exp_main} compares \Change, \Conf and \Rand on the test set.  We additionally evaluate the receiver-side criteria on the diagnostic subset (Supp.~\ref{sec:supp_diag}). At approximately 3\% of the native communication cost, \Change improves \apstrict by 9.0 points over the delayed native message, compared with 1.9 for \EgoTemp, 1.1 for JEPA error, 0.3 for \Rand and 0.1 for \Conf.  The difference reflects what each criterion captures: \Conf identifies where objects are, but a confidently detected stopped vehicle may already be represented correctly in the stale background.  Increasing the communication budget can even hurt when additional bytes introduce another frame of delay.  Specifically, a delta stream using 11.3\% of the native bytes arrives one frame late and scores 5.4 points below a stream using only 6.1\% of the bytes (95\% CI $[-6.90,-4.04]$; Supp.~\ref{sec:supp_diag}).

\subsection{Selection belongs to the sender}
\label{sec:asym}

The gap between sender-side and receiver-side criteria follows from the information available to each side. Consider a single link, a single frame, and cells of a single scale. If cell $k$ is refreshed, the receiver uses $\bz_k(t)$ and incurs zero error. Otherwise, it retains $\bh_k$ and incurs an error $d_k=d(\bz_k(t),\bh_k)\geq 0$. Every refreshed cell requires the same number of bytes, and the budget permits refreshing at most $n$ cells.  

\needspace{3\baselineskip}
\begin{proposition}[Information asymmetry]
\label{prop:asym}
Let $S^\star$ denote the set of  $n$ cells with the largest $d_k$. For every realization of the scene and every set $S$ with $|S|\le n$, including any set selected from the receiver's information alone, $\sum_{k\notin S^\star} d_k \le \sum_{k\notin S} d_k$. The sender can compute $S^\star$ because it observes $\bz_k(t)$ and can reconstruct $\bh_k$; the receiver cannot, because it does not observe $\bz_k(t)$.
\end{proposition}

The proposition is proved in Supp.~\ref{sec:supp_expected}. The best strategy available to a receiver is to rank cells by $\mathbb{E}[d_k\mid\text{receiver information}]$, which, under squared error, corresponds to the JEPA-error criterion defined above (Supp.~\ref{sec:supp_expected}). This ranking matches the sender's ranking only where the predictive variance is small, which is not the case for moving objects.  Consequently, a receiver-side model can improve performance only by reconstructing unsent cells more accurately than $\bh_k$, rather than by selecting them. Even this reconstruction-based benefit is unreliable without feedback. When the receiver fills the unsent cells using the predictor's output on the diagnostic subset, AP decreases by 6.7 points, despite an approximately 50\% reduction in latent error (Supp.~\ref{sec:supp_fill}). The receiver cannot distinguish accurate predictions from inaccurate ones, whereas the sender can. This information asymmetry motivates the closed loop design introduced subsequently in  \cref{sec:method}.

\section{\ours: Closed-Loop Latent Prediction}
\label{sec:method}

\Cref{prop:asym} assigns selection to the sender and leaves prediction to the receiver. \ours integrates these two roles into a unified loop (\cref{fig:method}). Both ends execute the same latent predictor $W$. The sender maintains an exact mirror of the receiver's state and verifies each candidate reconstruction against its current feature. It transmits feature data only for cells that cannot be adequately reconstructed by any available candidate, thereby avoiding unnecessary feature transmission while preserving reconstruction fidelity.

\subsection{Receiver state and its mirror}
\label{sec:mirror}

For a link $j\to i$, the receiver state $\M(t)$ stores a feature for every BEV cell $k$ at three scales. A cell's value may originate from a keyframe, the most recent increment that updated it, or the predictor; alternatively, the cell may be cleared to zero. Given the messages and their arrival frames, the poses, and the predictor $W$, the receiver-state update is deterministic. The sender has access to all the information needed to reproduce this update: it generates the messages, the link model determines their delays, it receives the request maps, and the poses are shared (\cref{sec:setting}).  The sender can therefore maintain a mirror $\hM(t)$ by applying the same update rule. Across the 38,810 frames and 95,446 link-frames of the test replay, the mirror reproduces the receiver state exactly, with zero error.

\subsection{Shared latent predictor}
\label{sec:predictor}

\def\wc{0.17}%
\newcommand{\wgrid}[3]{%
  \fill[white] (#2,#3) rectangle ++(6*\wc,3*\wc);
  \draw[black!13, line width=0.25pt, xshift=#2 cm, yshift=#3 cm, step=\wc] (0,0) grid (6*\wc,3*\wc);
  \draw[black!60, line width=0.5pt] (#2,#3) rectangle ++(6*\wc,3*\wc);
  \coordinate (#1-sw) at (#2,#3);
}
\newcommand{\wcell}[4][mjfeat]{\fill[#1] ($(#2-sw)+({(#3-1)*\wc+0.014},{(#4-1)*\wc+0.014})$) rectangle ++(\wc-0.028,\wc-0.028);}
\begin{figure}[!htbp]
\centering
\makebox[\linewidth][c]{%
\begin{tikzpicture}[x=1cm,y=1cm, font=\scriptsize, >={Stealth[length=1.5mm,width=1.3mm]},
  lbl/.style={font=\scriptsize, inner sep=1pt},
  flow/.style={->, line width=0.6pt, black!75}]
\fill[black!5, rounded corners=3pt] (0,1.68) rectangle (8.22,3.60);
\fill[mjsl, rounded corners=3pt] (0,0) rectangle (8.22,0.88);
\node[rotate=90, lbl, align=center] at (0.22,2.64) {\textbf{both ends,}\\[-2pt]\textbf{same inputs}};
\node[rotate=90, lbl] at (0.22,0.44) {\textbf{target}};
\foreach \i in {3,2,1}{
  \fill[white] ({0.52+0.04*\i},{2.42+0.04*\i}) rectangle ++(6*\wc,3*\wc);
  \draw[black!45, line width=0.4pt] ({0.52+0.04*\i},{2.42+0.04*\i}) rectangle ++(6*\wc,3*\wc);
}
\wgrid{h}{0.52}{2.42}
\wcell{h}{1}{1} \wcell{h}{2}{1} \wcell{h}{2}{3} \wcell{h}{3}{3}
\node[lbl, anchor=west] at (0.48,3.47) {held states $\M(t{-}1),\dots,\M(t{-}4)$,};
\node[lbl, anchor=west] at (0.48,3.24) {warped to the current ego frame};
\draw[flow] (1.70,2.68) -- (1.98,2.68);
\draw[black!70, line width=0.6pt, rounded corners=2pt, fill=white] (1.98,2.38) rectangle (3.98,2.98);
\node[inner sep=0pt, align=center, font=\scriptsize] at (2.98,2.68) {{\small $W$}\\[-2.5pt]0.65\,M parameters};
\node[lbl, anchor=west, text=mjmute] at (0.48,2.10) {age $\mathbf{a}$, support $\mathbf{s}$,};
\node[lbl, anchor=west, text=mjmute] (aux) at (0.48,1.87) {poses $\mathbf{T}_{1:4}$, sender type $c$};
\draw[->, line width=0.5pt, mjmute] (2.36,2.10) -- (2.62,2.10) -- (2.62,2.38);
\draw[flow] (3.98,2.68) -- (4.26,2.68);
\wgrid{d}{4.26}{2.42}
\draw[->, mjpred, line width=0.8pt, >={Stealth[length=1.1mm,width=1.0mm]}] ($(d-sw)+(1.5*\wc,2.5*\wc)$) -- ($(d-sw)+(3.45*\wc,1.55*\wc)$);
\draw[->, mjpred, line width=0.8pt, >={Stealth[length=1.1mm,width=1.0mm]}] ($(d-sw)+(2.5*\wc,2.5*\wc)$) -- ($(d-sw)+(4.45*\wc,1.55*\wc)$);
\node[lbl, anchor=west] at (4.60,3.47) {displacement field,};
\node[lbl, anchor=west] at (4.38,3.24) {$\le$\,20\,m/s $\times$ age};
\draw[flow] (5.28,2.68) -- (5.56,2.68);
\draw[black!70, line width=0.6pt, rounded corners=2pt, fill=white] (5.56,2.38) rectangle (6.60,2.98);
\node[inner sep=0pt, align=center, font=\scriptsize] at (6.08,2.68) {transport\\[-2pt](bilinear)};
\fill[black!75] (1.83,2.68) circle (0.03);
\draw[->, line width=0.6pt, black!75, rounded corners=1.5pt] (1.83,2.68) -- (1.83,3.07) -- (6.38,3.07) -- (6.38,2.98);
\draw[flow] (6.60,2.68) -- (7.12,2.68);
\wgrid{p}{7.12}{2.42}
\wcell{p}{1}{1} \wcell{p}{2}{1} \wcell[mjfeat!42]{p}{4}{2} \wcell[mjfeat!42]{p}{5}{2}
\draw[mjpred, line width=0.6pt] ($(p-sw)+(3*\wc+0.012,1*\wc+0.012)$) rectangle ++(2*\wc-0.024,\wc-0.024);
\node[lbl, anchor=east] at (8.18,3.24) {prediction $\bw(t)$};
\node[lbl, anchor=east, text=mjmute] at (7.22,2.10) {moves held features; adds none};
\foreach \x/\y in {0.60/0.30,0.70/0.52,0.78/0.38,0.88/0.60,0.95/0.27,1.02/0.47,1.10/0.62,0.64/0.63,0.84/0.24,1.12/0.34}{\fill[scptc] (\x,\y) circle (0.028);}
\node[lbl, anchor=west, align=left] at (1.22,0.44) {collaborator's\\[-2pt]current sweep};
\draw[flow] (2.78,0.44) -- (3.15,0.44);
\draw[black!70, line width=0.6pt, rounded corners=2pt, fill=white] (3.15,0.17) rectangle (5.55,0.71);
\node[lbl] at (4.35,0.44) {frozen CoSDH encoder};
\draw[flow] (5.55,0.44) -- (7.12,0.44);
\node[lbl, anchor=south] at (6.32,0.46) {$\bz_j(t)$};
\wgrid{z}{7.12}{0.185}
\wcell{z}{1}{1} \wcell{z}{2}{1} \wcell{z}{4}{2} \wcell{z}{5}{2}
\draw[<->, line width=0.6pt, black!75, densely dashed] (7.42,2.40) -- (7.42,0.72);
\draw[<->, line width=0.6pt, black!75] (7.86,2.40) -- (7.86,0.72);
\draw[line width=0.6pt, black!75, densely dashed] (1.20,1.48) -- (1.66,1.48);
\node[lbl, anchor=west] at (1.74,1.48) {\textbf{training:} loss in latent space, no labels};
\draw[line width=0.6pt, black!75] (1.20,1.24) -- (1.66,1.24);
\node[lbl, anchor=west] at (1.74,1.24) {\textbf{protocol:} the sender has $\bz_j(t)$ and verifies};
\node[lbl, anchor=west] at (1.74,1.01) {$\bw(t)$ cell by cell (Sec.~\ref{sec:modes})};
\end{tikzpicture}}
\caption{\textbf{Shared predictor $W$.} Both ends transport the same held states along a bounded displacement field; $W$ introduces no new content. It is trained without labels to predict the frozen encoder's feature of the collaborator's current sweep, and the sender verifies each predicted cell against the corresponding feature (\cref{sec:modes}; Supp.~\ref{sec:supp_train}).}
\label{fig:predictor}
\end{figure}
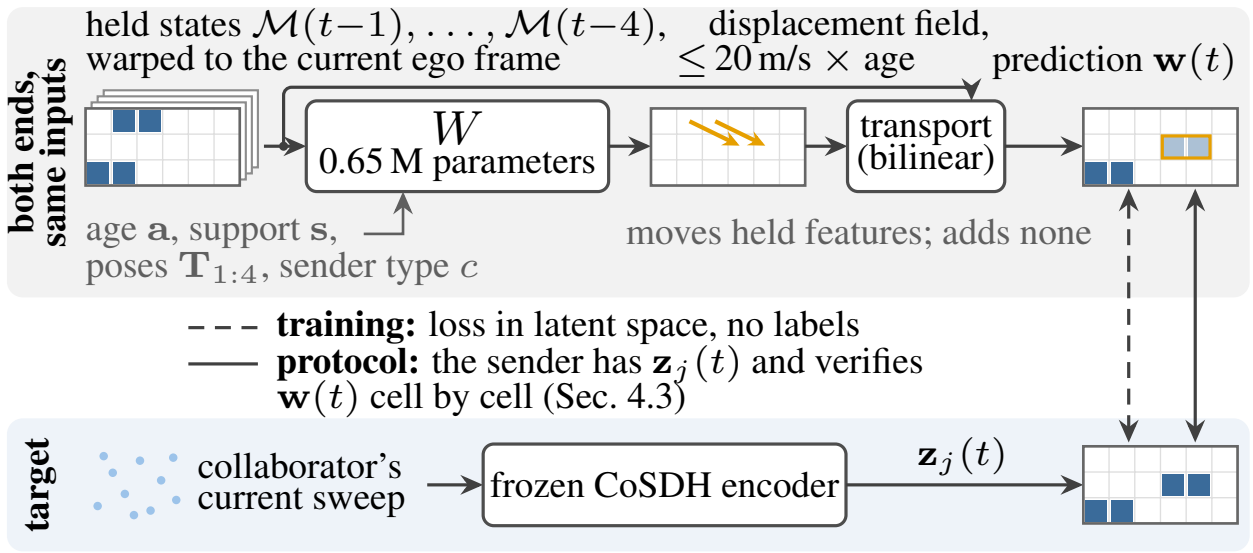

The predictor $W$ (\cref{fig:predictor}) takes as input, for each link, the last four end-of-frame states of the mirror, warped into the current ego frame, together with the age of each cell's feature $\mathbf{a}$, a support mask $\mathbf{s}$, the four relative poses $\mathbf{T}_{1:4}$ and the sender type $c$:
\begin{equation}
\bw(t) = W\big(\M(t{-}1),\dots,\M(t{-}4);\ \mathbf{a},\,\mathbf{s},\,\mathbf{T}_{1:4},\,c\big).
\label{eq:predictor}
\end{equation}
At three scales (0.8, 1.6 and 3.2\,m per cell) it outputs one displacement field per history slot, bounded by 20\,m/s multiplied by the age of the cell, and transports the held features along this field by bilinear sampling. The model has 0.65\,M parameters, and the protocol uses only the transported features. All inputs other than the current pose are fixed at the end of the previous frame. Both ends can therefore evaluate $W$ as soon as the pose is available, before the increment arrives.

$W$ is trained without labels in the latent space of the frozen CoSDH encoder. The prediction target is the encoder's output for the collaborator's current LiDAR sweep, and the loss is computed between representations in the latent space rather than on the original input space. Because the encoder is frozen, it also serves as the target encoder. In this sense, we refer to $W$ as a JEPA-style predictor (\cref{sec:related}; Supp.~\ref{sec:supp_train}).

\subsection{Verified modes and cell selection}
\label{sec:modes}

At frame $t$ the receiver can form three reconstructions of a cell without new data: retaining the feature that it holds, using the prediction, or clearing the cell to zero. The sender forms the same three reconstructions on its mirror. Let $d^h$, $d^w$ and $d^0$ denote the mean absolute difference over channels between the sender's current feature and, respectively, the held feature, the predicted feature, and zero. The candidate cells are those that the predictor supports or that an earlier increment, prediction, or clear operation has modified. On these cells, the sender marks a cell \Pred if $d^w<0.9\min(d^h,d^0)$, \Clear if $d^0<0.9\min(d^h,d^w)$, and \Hold otherwise. It then adds \Send cells from among those that the receiver requested and the sender observes, in decreasing order of remaining error, until the byte budget is reached. A \Send cell requires 12, 20 or 36 bytes at the three scales. The \Pred and \Clear marks are coded as a mode map, using either a sparse or a zlib-compressed representation, whichever is shorter, and its bytes count toward the budget.

The prediction must operate within this loop. On the diagnostic subset, the transported features yield a lower error than holding on 5.1\% of the candidate cells (Supp.~\ref{sec:supp_fill}). The sender's mirror identifies these cells and marks only them.

\needspace{3\baselineskip}
\begin{proposition}[No worse than holding, per cell]
\label{prop:closed}
Let $e_k$ denote the feature distance of the mode selected for an unsent candidate cell $k$. Then $e_k\le d^h_k$, with strict inequality if the cell is marked \Pred or \Clear. Hence, for any fixed set of sent cells, the summed feature distance of the unsent cells is at most that of the same loop with every unsent cell held.
\end{proposition}

\subsection{Keyframe mode}
\label{sec:keyframe}

The keyframe protocols transmit one native message every ten frames , which is used eight frames after capture, as is any native message. Between keyframes, they transmit increments at $r=4.4\%$, which the link model delivers within the frame. The increments never exceed 6.4\,KB per receiver per frame. With the keyframe averaged over its cycle, the downlink to a receiver is 15.3\,KB per frame on the test set, compared with the 20\,KB that the link carries.

We compare three keyframe protocols at equal bytes. \ChangeKF maintains no state between keyframes. The receiver uses the latest keyframe that has arrived together with the most recent increment, the sender ranks cells by their change relative to that keyframe, and no \Clear mode is available. This protocol serves as our CoST-style baseline~\cite{tang2025cost}. \ChangeM adds the mirror: increments accumulate in the receiver state, and each unsent cell is either held or cleared, but never predicted. \ours additionally includes the predictor. All three protocols are causal: keyframes are at least eight frames old, and request maps are taken from the frame preceding capture.

\section{Experiments}
\label{sec:exp}

The detector remains frozen throughout, and only $W$ is trained. Each baseline isolates one published idea (Where2comm~\cite{hu2022where2comm} for \Conf, CooperTrim~\cite{mukhopadhyay2026coopertrim} for \EgoTemp, and CoST~\cite{tang2025cost} for \ChangeKF) on the same frozen detector, link, and byte budget, so that the conditions differ only in how cells are selected and updated. The replay operates on features that have passed through CoSDH's compression and its decoder, and bytes are counted in the compressed format. The keyframe protocols are evaluated on the test set with the configuration and byte target that were frozen on DEV. Results on the test set follow criteria established before  evaluation (Supp.~\ref{sec:supp_hygiene}). Three analyzes were not registered: the comparisons of the keyframe protocols with the conditions of \cref{tab:main} (marked $^\S$ in \cref{tab:ladder}), the robustness checks of Supp.~\ref{sec:supp_robust}, and the object-level counts of ghosts and localized objects.

\begin{table}[!htbp]
\caption{\textbf{Which cells to refresh, test set} (3,936 queries, every-frame protocol). $r$: delta downlink bytes relative to the native message; $D$: delay in frames of the delta, or of the native message in the \Native rows. Rec.: share of the delay loss recovered (\%). $^\dagger$Not achievable on the measured link.}
\label{tab:main}
\centering
\footnotesize
\setlength{\tabcolsep}{2.6pt}
\begin{adjustbox}{max width=\linewidth}
\begin{tabular}{@{}lrrcrcr@{}}
\toprule
& & & \multicolumn{2}{c}{BEV AP@0.7} & \multicolumn{2}{c}{3D AP@0.7} \\
\cmidrule(lr){4-5}\cmidrule(lr){6-7}
Method & $r$ (\%) & $D$ & AP & Rec. & AP & Rec. \\
\midrule
\textcolor{black!55}{\Native, delay-free$^\dagger$} & \textcolor{black!55}{--} & \textcolor{black!55}{0} & \textcolor{black!55}{46.96} & \textcolor{black!55}{100} & \textcolor{black!55}{23.78} & \textcolor{black!55}{100} \\
\Native (C-V2X) & 0 & 8 & 34.20 & 0 & 16.99 & 0 \\
\midrule
+ \Rand & 5.79 & 0 & 35.13 & 7.3 & 17.53 & 8.0 \\
+ \Conf~\cite{hu2022where2comm} & 3.89 & 0 & 34.77 & 4.5 & 17.25 & 3.9 \\
+ \Change ($\tau$=0.98) & 2.46 & 0 & 43.49 & 72.8 & 21.83 & 71.3 \\
+ \Change ($\tau$=0.95) & 5.79 & 0 & \textbf{44.58} & \textbf{81.4} & \textbf{22.55} & \textbf{81.9} \\
\bottomrule
\end{tabular}
\end{adjustbox}
\end{table}

\subsection{Which cells to refresh}
\label{sec:exp_main}

On the test set (\cref{tab:main}), transmitting the \Change delta at $\tau=0.95$ recovers 81.9\% of the AP lost to the eight-frame delay under \apstrict, and 78.5--81.4\% under BEV AP@0.5, BEV AP@0.7 and 3D AP@0.5 (Supp.~\ref{sec:supp_tables}). Confidence-ranked and random cell selection recover at most 8.1\% under any of the four metrics. With 20\% packet loss, \Change still recovers 62.5--64.7\%. This finding holds in each of the two test scenes (\Change 77--83\%, \Conf at most 5.3\%, \Rand at most 10.4\%) and when 5-second blocks of queries are resampled (Supp.~\ref{sec:supp_robust}).

\subsection{Mirror-JEPA at equal bytes}
\label{sec:exp_ladder}

\begin{table*}[!htbp]
\caption{\textbf{Keyframe mode at equal bytes, test set} (3,936 queries, one keyframe every ten frames; the increments of the three protocols are 4.4\% of the native downlink and equal to within 0.1\%). Gray: the native message streamed every frame with the \Change delta (best row of \cref{tab:main}). Paired differences with 95\% CIs; bold: CI excludes 0. Bytes: per link-frame. $^\ast$Preregistered primary metric. $^\S$Not registered.}
\label{tab:ladder}
\centering
\footnotesize
\setlength{\tabcolsep}{5pt}
\begin{adjustbox}{max width=\linewidth}
\begin{tabular}{@{}lccccrr@{}}
\toprule
& BEV AP@0.5 & BEV AP@0.7 & 3D AP@0.5 & 3D AP@0.7$^\ast$ & Down (KB) & Up (KB) \\
\midrule
\textcolor{black!55}{Native every frame + \Change} & \textcolor{black!55}{66.87} & \textcolor{black!55}{44.58} & \textcolor{black!55}{52.98} & \textcolor{black!55}{22.55} & \textcolor{black!55}{45.7} & \textcolor{black!55}{14.6} \\
\ChangeKF (CoST-style~\cite{tang2025cost}) & 65.90 & 43.07 & 52.01 & 21.36 & 6.2 & 7.3 \\
\ChangeM & 68.58 & 44.74 & 55.25 & 21.87 & 6.2 & 7.3 \\
\textbf{\ours} & 68.88 & 45.29 & 55.98 & 21.90 & 6.2 & 7.3 \\
\addlinespace[2pt]
\ours $-$ \ChangeKF & \textbf{+2.98} \ci{2.72, 3.24} & \textbf{+2.21} \ci{1.93, 2.48} & \textbf{+3.97} \ci{3.66, 4.25} & \textbf{+0.54} \ci{0.30, 0.81} & & \\
\ChangeM $-$ \ChangeKF & \textbf{+2.68} \ci{2.40, 2.94} & \textbf{+1.67} \ci{1.39, 1.94} & \textbf{+3.24} \ci{2.95, 3.51} & \textbf{+0.51} \ci{0.27, 0.76} & & \\
\ours $-$ \ChangeM & \textbf{+0.30} \ci{0.17, 0.45} & \textbf{+0.54} \ci{0.38, 0.71} & \textbf{+0.73} \ci{0.57, 0.90} & +0.03 \ci{$-$0.14, 0.19} & & \\
\ours $-$ every frame$^\S$ & \textbf{+2.01} \ci{1.76, 2.27} & \textbf{+0.70} \ci{0.44, 0.99} & \textbf{+3.00} \ci{2.68, 3.31} & \textbf{$-$0.65} \ci{$-$0.91, $-$0.38} & & \\
\bottomrule
\end{tabular}
\end{adjustbox}
\end{table*}

\begin{figure*}[!htbp]
\centering
\includegraphics[width=\textwidth]{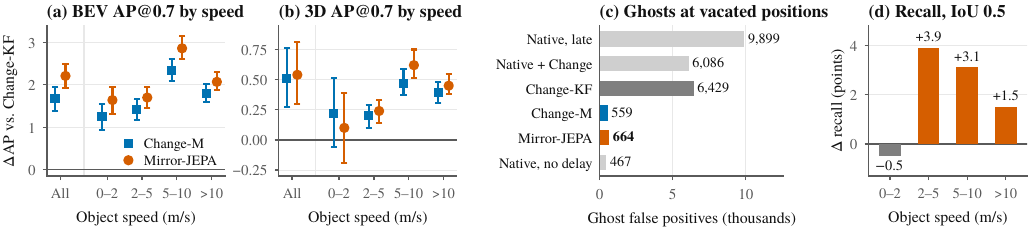}
\caption{\textbf{Sources of the improvement} (test set). (a, b) Paired AP differences relative to \ChangeKF by object speed, with 95\% CIs; ``All'' denotes the overall difference. The bins contain 50,555, 5,200, 8,944 and 4,039 objects. (c) False positives at positions that a vehicle vacated within the previous 1.7\,s (ghosts), at IoU 0.5. Light gray: conditions of \cref{tab:main}. (d) Change in recall from \ChangeKF to \ours.}
\label{fig:gain}
\end{figure*}

On the held-out test set (\cref{tab:ladder}), \ours outperforms \ChangeKF at equal bytes under all four metrics: +2.98 BEV AP@0.5, +2.21 BEV AP@0.7, +3.97 3D AP@0.5 and +0.54 \apstrict, each with a 95\% CI above zero. All registered criteria, with \apstrict as the primary metric, are satisfied (Supp.~\ref{sec:supp_hygiene}). Under \apstrict, the improvement is concentrated on moving objects (\cref{fig:gain}a,b): +0.24, +0.62 and +0.45 points in the 2--5, 5--10 and $>$10\,m/s bins, all with CIs above zero, compared with +0.10 ($[-0.19,0.39]$) below 2\,m/s. Under BEV AP@0.5, the improvement is 8.1 to 11.1 points for moving objects and 2.0 points for near-static objects.

\begin{figure*}[!htbp]
\centering
\includegraphics[width=\linewidth]{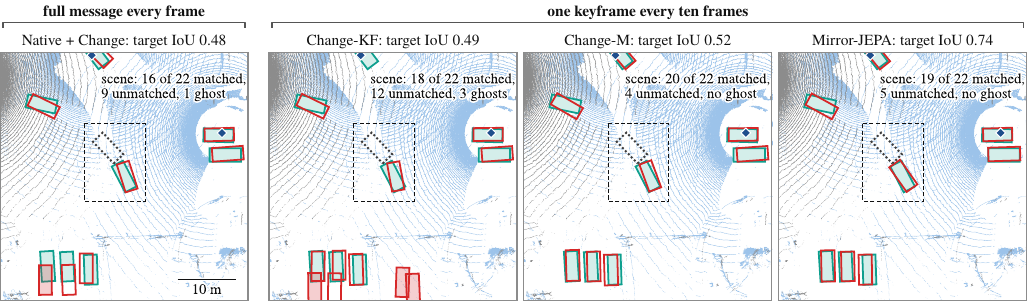}
\caption{\textbf{The query of \cref{fig:method} under the four conditions that deliver fresh content}, viewed as in \cref{fig:method}a. Titles: best 3D IoU with the target (dashed frame). The every-frame stream and \ChangeKF remain below 0.5 and leave ghosts (shaded); \ChangeM leaves none, and \ours fits the target more closely. The target was selected among objects that \ChangeKF misses and \ours localizes (Supp.~\ref{sec:supp_qual}).}
\label{fig:closed}
\end{figure*}

\paragraph{Closed loop and predictor.}
Most of the improvement originates from the closed loop (\cref{fig:closed}; Supp.~Fig.~\ref{fig:supp_qual2}). Among test objects moving at 2--10\,m/s that are not cooperating vehicles, 137 change from an IoU below 0.5 under \ChangeKF to at least 0.7 under \ours and 43 change in the opposite direction (Supp.~\ref{sec:supp_robust}). \ChangeM, which holds, refreshes and clears cells but does not predict, already improves on \ChangeKF under all four metrics, by 1.67 BEV AP@0.7 and 0.51 \apstrict. By point estimate, this corresponds to 75--94\% of the improvement of \ours, depending on the metric. The predictor adds a further increment at equal bytes: +0.30 BEV AP@0.5, +0.54 BEV AP@0.7 and +0.73 3D AP@0.5, all with CIs above zero. \ours transmits 15.1\% fewer feature cells than \ChangeM and instead marks 2.0 million cells for transport, approximately ten for each feature cell that it forgoes (Supp.~\ref{sec:supp_bytes}). The contribution of the predictor depends on object motion. In the three moving-speed bins it is positive with a CI above zero in 11 of the 12 metric-bin pairs, including \apstrict at 5--10\,m/s (+0.14, $[0.07,0.22]$) and above 10\,m/s (+0.06, $[0.01,0.12]$). For near-static objects, which account for 72\% of the ground-truth boxes, it is indistinguishable from zero under \apstrict ($-0.12$, $[-0.33,0.07]$), and the overall \apstrict changes only marginally (+0.03, $[-0.14,0.19]$).

\paragraph{Confirmation on DEV.}
On the DEV set, which serves as the confirmation set, \ours also outperforms \ChangeKF under all four metrics (+0.59 \apstrict, $[0.29,0.91]$). The predictor's increment on DEV has a CI above zero under \apstrict (+0.33, $[0.12,0.54]$). All registered DEV outcomes, including one criterion that was not met, are reported in Supp.~\ref{sec:supp_dev} and~\ref{sec:supp_hygiene}.

\paragraph{Robustness.}
The test set comprises two scenes of consecutive frames. In each scene \ours outperforms \ChangeKF under BEV AP@0.5, BEV AP@0.7 and 3D AP@0.5 with CIs above zero (+1.9 to +4.1). These three improvements, as well as the predictor's increment over \ChangeM under the same metrics, retain CIs above zero when 5-second blocks of queries are resampled. The improvement under \apstrict is less robust: it originates from one scene (+2.45 in scene 8 compared with $-0.06$ in scene 2), with a block-bootstrap CI of $[-0.08,1.27]$. For objects moving faster than 2\,m/s, the improvement has a CI above zero in both scenes under all four metrics (Supp.~\ref{sec:supp_robust}).

\subsection{Sources of the improvement}
\label{sec:exp_gain}

Two effects underlie the improvement over \ChangeKF. The first is a reduction in stale ghosts (\cref{fig:gain}c). On the test set, \ChangeKF produces 6,429 false positives that lie within 2\,m of a position that a vehicle occupied during the previous 1.7\,s and has since left, whereas \ChangeM and \ours produce 559 and 664, respectively. Overall, \ours produces 24\% fewer false positives at IoU 0.5 (29,869 compared with 39,529). The second effect is that more moving objects are detected (\cref{fig:gain}d): at IoU 0.5, recall increases by 1.5 to 3.9 points in the three moving-speed bins and decreases by 0.5 points for near-static objects. Because every unmatched prediction counts as a false positive in each speed bin, fewer false positives increase AP in all bins.

The same pattern holds relative to the full message itself (\cref{fig:gain}c, light gray). The the late message leaves 9,899 false positives at vacated positions and the every-frame stream leaves 6,086, compared with 664 for \ours and 467 without delay.

\paragraph{Cooperating vehicles as targets.}
The ground truth includes the cooperating vehicles themselves (11.5\% of the test objects), which could report their own pose. When their boxes are ignored, the improvement over \ChangeKF is +2.99, +1.94, +3.95 and +0.45 under the four metrics, all with CIs above zero, compared with +2.98, +2.21, +3.97 and +0.54 when they are included. The localization and ghost counts change only slightly (Supp.~\ref{sec:supp_robust}).

\subsection{Keyframe mode versus every-frame streaming}
\label{sec:exp_bytes}

On the test set, we also compare with streaming the full native message every frame, which corresponds to the \Change condition of \cref{tab:main} (re-evaluated with the keyframe implementation; Supp.~\ref{sec:supp_hygiene}). Using about 4.5 times fewer bytes (13.5 compared with 60.3\,KB per link-frame in both directions), \ours scores higher under BEV AP@0.5 (+2.01), BEV AP@0.7 (+0.70) and 3D AP@0.5 (+3.00), and lower under \apstrict ($-0.65$); all four CIs exclude zero (\cref{tab:ladder}). Under block resampling,  the BEV AP@0.7 interval includes zero ($[-0.03,1.36]$), whereas the other three differences retain keep their sign (per-scene results in Supp.~\ref{sec:supp_robust}). On DEV the same comparison yields +0.24, $-0.79$, $-0.21$ and $-0.78$. Relative to the native CoSDH message over C-V2X, \ours improves BEV AP@0.5 by 14.5 points, BEV AP@0.7 by 11.1 points and \apstrict by 4.9 points with one-seventh of the downlink (\cref{tab:benchmark}; point estimates).

\paragraph{Computational cost.}
On an otherwise idle NVIDIA L40S, the predictor together with its warp requires 8.8\,ms per link and can be executed before the increment arrives, and the receiver's update requires 2.6\,ms. Encoding the point clouds of one query requires 17.9\,ms (medians; Supp.~\ref{sec:supp_latency}). Our reference implementation of cell selection requires 126\,ms per link-frame, which exceeds the 100\,ms frame interval. It re-encodes the entire mode map after each candidate cell, an overhead that incremental tracking of the encoded size would largely avoid.

\section{Conclusion}
\label{sec:conclusion}

Over a real C-V2X link, much of the perception accuracy lost to communication delay can be recovered by transmitting changes relative to the receiver's current feature state. The proposed sender-side state mirror gives the sender an exact representation of the receiver's state, enabling the sender to select cell-level updates and verify predictions from a shared latent predictor before they are used by the receiver. With one keyframe every ten frames, \ours outperforms a causal keyframe baseline at equal bytes on the test set and passes every criterion fixed beforehand. Relative to the full CoSDH messages on the same link, it gains 14.5 BEV AP@0.5 and 4.9 \apstrict with one-seventh of the downlink. Compared with streaming full messages every frame, it reduces communication by a factor of 4.5, achieves performance within 0.8 AP of the full-stream baseline across all four metrics, and reduces false positives at vacated vehicle positions by approximately 90\%. Importantly, the proposed closed-loop protocol is independent of the specific latent predictor. The predictor $W$ can be replaced with a more capable model without changing the underlying communication mechanism. This separation provides a flexible foundation for improving cooperative perception as prediction models advance and C-V2X communication constraints persist.

\paragraph{Limitations.}
Our link model follows the benchmark's latency accounting at the stream level. The native messages in our replay are smaller than the benchmark's (43.2 against 59.5\,KB per collaborator), and we keep its eight frames for every query. At the same throughput, our smaller messages would incur delays of approximately three frames with two collaborators and seven frames with three (Supp.~\ref{sec:supp_protocol}). Several additional factors are not captured in the replay. First, it accounts only for transmission delay and excludes computation time. Second, it assumes zero-delay pose exchange. Third, it does not model contention between keyframes and incremental updates. When averaged over a keyframe cycle, their combined downlink exceeds the link's 20\,KB per frame in 34\% of receiver frames.  Our communication accounting also has a limitation related to the feature decoder. We count bytes per transmitted cell, whereas CoSDH's convolutional decoder associates each decoded cell with a $5\times5$  neighborhood of latent features. Transmitting these neighborhoods would increase the communication cost of the delta of \cref{tab:main} by approximately sixfold, exceeding what the link can deliver within a single frame (Supp.~\ref{sec:supp_protocol}).

Regarding the predictor, we do not have a matched model trained without the JEPA objective, preventing us from isolating the contribution of JEPA-based training from other factors in the predictor's performance.  Our experiments also use a single frozen detector and one dataset, the only dataset we are aware of that reports measured C-V2X throughput. The held-out test set contains two scenes and 3,936 queries, which we analyze separately in \cref{sec:exp_ladder}.  Finally, we do not compare with retrained systems, including the latency-compensation methods of \cref{sec:related} or evaluate CoST as a whole. Nor do we evaluate the keyframe protocols under packet loss. These scenarios remain important directions for assessing the protocol's performance under more diverse and challenging deployment conditions.

\section{Acknowledgements}
This research was supported by the U.S. Department of Transportation, Office of the Assistant Secretary for Research and Technology (OST-R), University Transportation Centers Program, through the Center for Regional and Rural Connected Communities (CR2C2) under Grant No. 69A3552348304.

\section{LLM Usage Statement}

We used two large language model (LLM) tools in this work: Claude (Anthropic) and Codex (OpenAI). These tools assisted with manuscript preparation, code implementation, and the generation of plots and figures. All generated content and outputs were subsequently reviewed, revised as needed, and approved by the authors. The research ideas, methodology, and conclusions are those of the authors, who take full responsibility for the accuracy and integrity of the content presented in this paper.

\bibliographystyle{unsrt}
\bibliography{main}

\clearpage
\begin{center}
{\LARGE\sc Supplementary Material\par}
\end{center}
\appendix
\crefalias{section}{appendix}\crefalias{subsection}{subappendix}
\setcounter{table}{0}\renewcommand{\thetable}{S\arabic{table}}\renewcommand{\theHtable}{S\arabic{table}}
\setcounter{figure}{0}\renewcommand{\thefigure}{S\arabic{figure}}\renewcommand{\theHfigure}{S\arabic{figure}}
\setcounter{equation}{0}\renewcommand{\theequation}{S\arabic{equation}}\renewcommand{\theHequation}{S\arabic{equation}}
\renewcommand{\topfraction}{0.92}\renewcommand{\dbltopfraction}{0.92}\renewcommand{\textfraction}{0.06}
\renewcommand{\floatpagefraction}{0.82}\renewcommand{\dblfloatpagefraction}{0.82}\setcounter{topnumber}{4}\setcounter{dbltopnumber}{4}\setcounter{totalnumber}{6}

\paragraph{Overview.}
Section~\ref{sec:supp_protocol} specifies the link model and the byte accounting behind Sec.~\ref{sec:setting} and the Limitations. Sections~\ref{sec:supp_expected}--\ref{sec:supp_diag} give the proofs and the diagnostic-subset evidence for Sec.~\ref{sec:analysis}. Section~\ref{sec:supp_train} describes the predictor. Section~\ref{sec:supp_tables} holds the complete tables and the qualitative examples for Sec.~\ref{sec:exp}, and Section~\ref{sec:supp_hygiene} the registered criteria with their outcomes, including those that failed, and the robustness checks.

\section{Link model and protocol details}
\label{sec:supp_protocol}

\paragraph{Delay.}
A downlink stream is delayed by
\begin{equation}
D=\left\lfloor \frac{0.1785\ \text{MB}\cdot r\cdot 8}{1.6\ \text{Mbps}}\cdot 10\right\rfloor=\lfloor 8.925\,r\rfloor\ \text{frames},
\end{equation}
where 0.1785\,MB and 1.6\,Mbps are the benchmark's message size and throughput for three collaborators, and $r$ is the ratio of the condition's mean downlink bytes to the mean native downlink bytes. $D$ is set once per condition, not per message. Native messages and keyframes have $D=8$. The six delta conditions of \cref{tab:main} and \cref{tab:supp_test} and the increments of the keyframe protocols have $D=0$; for each, the replay ran with $D=0$ and the bytes of the full split were then checked against the formula. A request of $b$ bytes generated in frame $g$ is usable in frame $g+1+\lfloor b/20{,}000\rfloor$. The added frame reflects that a request cannot be used in the frame that produces it. The delays contain no measured computation time and no queueing.

\paragraph{Sensitivity to the delay.}
\Cref{tab:supp_delay} gives the AP of the native message against its delay on DEV. The first frame of delay costs about half of the total loss only under 3D AP@0.7; under BEV AP@0.5 it costs 1.5 of 17.8 points on DEV. In its supplement, CoSDH reports the highest accuracy among the compared methods at a latency of 100\,ms on OPV2V~\cite{xu2025cosdh}. On this benchmark one frame costs little under the loose metric and about half of the total loss under the strict one.

\begin{table}[!htbp]
\caption{\textbf{AP of the native message against its delay} on DEV (2,406 queries).}
\label{tab:supp_delay}
\centering
\footnotesize
\setlength{\tabcolsep}{5pt}
\begin{adjustbox}{max width=\linewidth}
\begin{tabular}{@{}ccccc@{}}
\toprule
Delay & \multicolumn{2}{c}{BEV AP} & \multicolumn{2}{c}{3D AP} \\
\cmidrule(lr){2-3}\cmidrule(lr){4-5}
(frames) & @0.5 & @0.7 & @0.5 & @0.7 \\
\midrule
0 & 64.17 & 47.51 & 53.91 & 21.61 \\
1 & 62.67 & 42.24 & 51.33 & 17.81 \\
2 & 57.21 & 37.60 & 46.25 & 16.43 \\
3 & 52.90 & 35.99 & 43.00 & 16.04 \\
4 & 50.29 & 35.19 & 40.82 & 15.65 \\
5 & 48.59 & 34.65 & 39.72 & 15.36 \\
8 & 46.34 & 33.13 & 38.23 & 14.53 \\
\bottomrule
\end{tabular}
\end{adjustbox}
\end{table}

\paragraph{Byte accounting.}
We count bytes with 1\,KB $=$ 1,000\,B. A transmitted cell costs 12, 20 or 36 bytes at the three scales: 4, 8 or 16 compressed channels of two bytes each and a 4-byte sparse index. Every increment carries a 219-byte header. The ratio $r$ of a condition is its total delta downlink on a split divided by the total native downlink of that split. It includes mode maps and headers and excludes keyframes, the background and all uplink bytes. In keyframe mode, the downlink of a link-frame is its increment plus one tenth of the keyframe of its cycle, and the request uplink includes one tenth of the keyframe request. Request maps travel uplink as a run-length code.

\paragraph{Decoding sparse cells.}
CoSDH compresses each scale with a small convolutional autoencoder whose decoder has two $3\times3$ layers, so the decoded feature of a cell depends on the latents of a $5\times5$ neighborhood. Our replay charges a transmitted cell at the size of its own latent and applies the value that the cell has in the sender's fully decoded map; the native message and the keyframes are handled in the same way. Reproducing that value exactly at the receiver would require the neighborhood as well. For the \Change delta of \cref{tab:main}, sending the union of these neighborhoods would raise the delta from 2.5 to 15.5\,KB per link-frame, or from 5.8\% to 36\% of the native bytes, which the link model would no longer deliver within the frame. For the keyframe protocols we did not store the positions of the transmitted cells and cannot give the number. A receiver that merges the received latents with the ones it holds and then decodes the whole map needs no neighborhood and costs exactly the bytes we count, but it is not equivalent to our replay, and we have not evaluated it.

\paragraph{Mode map and candidate cells.}
In the keyframe protocols, the sender's current feature is its CoSDH feature after compression and decoding, expressed in the receiver's current frame and multiplied by the sender's supply mask. It is zero where the sender supplies nothing, which is how a held feature in a cell that the sender no longer supplies can come to be cleared. \Hold is the default and is not coded, and the positions of \Send cells travel with their features, so the mode map lists only \Pred and \Clear cells. It uses the shorter of a sparse list and a zlib-compressed map of two bits per candidate cell, behind a 1-byte tag. The candidate cells are the cells that the prediction supports, together with the cells that the mirror last updated with an increment, a prediction or a clear; the request map restricts which cells may be sent, separately from this set. \Send cells are added in decreasing order of remaining error across the three scales as long as the header, the payload and the mode map fit the budget; if the mode map alone exceeds the budget, the frame sends features only.

\paragraph{Load at the receiver.}
Summed over the links into a receiver, the increments of the keyframe protocols average 4.7\,KB per frame on the test set and never exceed 6.4\,KB. With one tenth of each keyframe added, the downlink averages 15.3\,KB (95th percentile 22.7\,KB, maximum 24.1\,KB), and 66.2\% of receiver frames stay within 20\,KB. The remaining 34\% are above 20\,KB, so a deployment would need a longer keyframe period for receivers with three collaborators. The replay sends each keyframe whole in one frame, sends no increment in that frame, and uses the keyframe eight frames later. The native messages into a receiver total 22.5\,KB with one collaborator (89 test queries), 69.6\,KB with two (1,665) and 143.8\,KB with three (2,087); 95 queries have no collaborator. Over all queries the average is 106\,KB per frame. Per collaborator, the native messages in our replay average 43.2\,KB, against the 59.5\,KB that the benchmark reports, and we keep the benchmark's eight frames for every query. The benchmark itself reports 298 and 595\,ms for CoSDH with one and two collaborators~\cite{wu2026cooperscene}. At 1.6\,Mbps the smaller messages of our replay would take about 110, 350 and 720\,ms, which is one, three and seven frames with one, two and three collaborators. On DEV the native message has lost 63--80\% of its eight-frame loss at three frames and 87--90\% at five, depending on the metric (\cref{tab:supp_delay}). For queries with fewer than three collaborators the eight-frame baseline is therefore pessimistic. The keyframes of all three keyframe protocols carry the same eight frames.

\paragraph{Every-frame protocol.}
All conditions of \cref{tab:main} share the same delayed background, the native message captured eight frames earlier. Our replay encodes this background with the receiver's request map at the current frame, not at capture time. Because every policy receives the same background, the comparison among policies is unaffected. Deltas are selected with the request map of the previous frame. The change score of \Change compares the current feature with the packet that the sender transmitted eight frames earlier. Under packet loss the sender keeps this reference, because it does not know which packets arrived. Forty-eight test queries (1.2\%) fall in the first eight frames of a sequence; the 32 of them that have collaborators receive a fresher background.

\paragraph{Keyframe protocols.}
Each keyframe protocol uses one constant per-link budget per frame, set to the target $r$ times the mean native downlink per link-frame of the split. A registered rule corrects the budget once, from bytes alone, if the realized $r$ deviates from the target by more than 5\%, and the protocol is then replayed. Both splits needed this correction (the first pass fell 9--11\% short). The realized $r$ is then 4.418\% for \ChangeKF and 4.421\% for \ChangeM and \ours on the test set, against a target of 4.421\%, and 4.40\% and 4.42\% on DEV. \ChangeKF uses the latest keyframe that has arrived and the most recent increment; in the frame in which a keyframe is sent, the previous increment stays in use. Cells outside the sender's current supply cannot be sent, so \ChangeKF has no means to erase a stale cell.

\paragraph{Detector.}
The detector is the \texttt{cosdh.pth} checkpoint of the benchmark's model zoo; the benchmark states that its released models are trained on the train set. Table~\ref{tab:benchmark} of the main paper reproduces the benchmark's own results for its six detectors with three collaborators~\cite{wu2026cooperscene}. CoSDH has the smallest messages, and it is the most accurate over C-V2X (mAP@0.7 of 0.41) although not without a bandwidth limit (0.52 against 0.57). For comparison, the CoSDH paper reports an AP@0.7 of 92.99 on the simulated OPV2V benchmark~\cite{xu2022opv2v} and 63.85 on real data, both without a bandwidth limit~\cite{xu2025cosdh}. On our test set the released checkpoint reaches a BEV AP@0.7 of 46.96 without delay and 34.20 at eight frames, and the two scenes differ widely (42.77 and 57.53 without delay; \cref{tab:supp_t1_scene}). The benchmark's values are not directly comparable with ours: its table does not say whether it uses BEV or 3D IoU, and the public release that we use has two test scenes.

\section{Proofs and receiver-side expected error}
\label{sec:supp_expected}

\paragraph{Proof of Proposition~\ref{prop:asym}.}
$\sum_{k\notin S} d_k=\sum_k d_k-\sum_{k\in S} d_k$, and since $d_k\ge0$, the $n$ largest terms maximize $\sum_{k\in S} d_k$ over $|S|\le n$. \qed

\paragraph{Proof of Proposition~\ref{prop:closed}.}
A cell is marked \Pred only if $d^w_k<0.9\,d^h_k$ and \Clear only if $d^0_k<0.9\,d^h_k$; otherwise it is held and $e_k=d^h_k$. \qed

\paragraph{Expected error at the receiver.}

Let the receiver's information be $\mathcal{I}$ and let $W$ give $P_k=\mathbb{E}[\bz_k\mid\mathcal{I}]$ and $\sigma_k^2=\mathbb{E}[\lVert\bz_k-P_k\rVert^2\mid\mathcal{I}]$. Then
\begin{equation}
\mathbb{E}\big[\lVert\bz_k-\bh_k\rVert^2\mid\mathcal{I}\big]=\lVert P_k-\bh_k\rVert^2+\sigma_k^2,
\end{equation}
since the cross term vanishes. This is the best receiver-side ranking score for the squared distortion. Ranking by it selects the same cells as the sender's \Change only if $\sigma_k^2$ is small relative to the spread of $\lVert\bz_k-\bh_k\rVert^2$ across cells.

\section{Why open-loop filling hurts}
\label{sec:supp_fill}

On the diagnostic subset, filling the unsent cells of the \Change delta with the predictor's output lowers 3D AP@0.7 by 6.7 points, from 40.84 to 34.12, at the same 6.1\% of the native bytes and without added delay. At a feature age of ten frames, holding gives a mean squared error of 0.088. The predictor gives 0.042 with its full context and 0.046 with the context input set to zero, which is how the fill experiment and the protocol run it. At ages of 5, 10 and 20 frames, the predictor reduces the error by 56--60\% outside ground-truth boxes and by 1--10\% inside them. These numbers use the same checkpoint as the keyframe protocols. Regressing with a squared loss averages over possible future positions of a moving object, which flattens the feature peaks that the frozen detection head responds to. Both observations motivate transport (\cref{sec:predictor}): moving observed features keeps their peaks, and the closed loop admits a prediction only where the sender has verified it against its current feature.

Inside the closed loop, the predictor is also not a safe replacement for the held feature on its own. Over the candidate cells of 339 link-frames of the diagnostic subset, the transported features carry 1.46 times the summed feature error of simply holding, and they beat holding on only 5.1\% of cells.

\section{Selection criteria on the diagnostic subset}
\label{sec:supp_diag}

\begin{figure}[!htbp]
\centering
\includegraphics[width=0.5\linewidth]{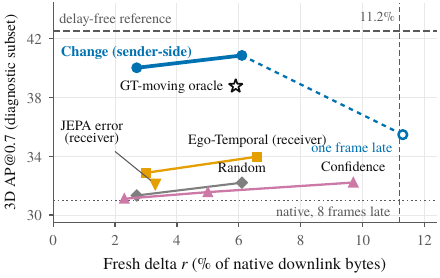}
\caption{\textbf{Which cells to refresh} (diagnostic subset of 128 queries). Sender-side \Change recovers most of the gap between the delayed native message and the delay-free reference at 3--6\% extra bytes. Receiver-side criteria, including the expected error of a JEPA-style world model, stay close to \Rand. A delta above 11.2\% of the native bytes is used one frame late. The subset comes from three training scenes of the benchmark, so only the comparison among conditions is meaningful. The receiver-side criteria, \Rand, the oracle and the fill come from a second run (\cref{sec:supp_diag}).}
\label{fig:selection}
\end{figure}

\Cref{fig:selection} compares the selection criteria of \cref{sec:select} on the diagnostic subset, and \cref{tab:supp_diag} gives the numbers behind it, with the \cooponly and \moving slices defined as in \cref{fig:teaser}a. At about 6\% of the native bytes, \Change adds 9.9 AP and a GT-moving oracle, which refreshes the cells of moving ground-truth boxes, adds 7.8. \Change also exceeds this oracle, which suggests that cells where new content enters the collaborator's view need refreshing too. With 2.7\% of the bytes \Change reaches 40.00, more than the 35.46 it reaches with 11.3\% one frame late. Each threshold is a quantile $\tau$ of the criterion's score, calibrated on these 128 queries. The rows marked $^\dagger$ come from a second run and are not a controlled comparison with the first; the caption states the differences.

\begin{table*}[!htbp]
\caption{Results on the diagnostic subset behind \cref{fig:selection}, 3D AP@0.7. S: sender-side, R: receiver-side. \moving: objects displaced by at least 0.5\,m over 0.5\,s. \cooponly: objects with fewer than five LiDAR points from the receiver and at least five from a collaborator. Rows marked $^\dagger$ come from a second run on the same queries. In that run the receiver-side criteria request cells by threshold alone, without the forced request for cells that have no history, and the fill row runs the predictor with zero context. They are diagnostic and not a controlled comparison with the first run. Dashes: not stored.}
\label{tab:supp_diag}
\centering
\footnotesize
\setlength{\tabcolsep}{6pt}
\begin{adjustbox}{max width=\linewidth}
\begin{tabular}{@{}llrcccc@{}}
\toprule
Criterion & Side & $r$ (\%) & $D_r$ & Overall & \cooponly & \moving \\
\midrule
\Native & -- & -- & 8 & 30.98 & 18.05 & 7.65 \\
\Change $\tau$=0.98 & S & 2.7 & 0 & 40.00 & 29.12 & 19.59 \\
\Change $\tau$=0.95 & S & 6.1 & 0 & 40.84 & 30.10 & 20.66 \\
\Change $\tau$=0.90 & S & 11.3 & 1 & 35.46 & 23.30 & 12.71 \\
\Conf $\tau$=0.98~\cite{hu2022where2comm} & S & 2.3 & 0 & 31.11 & 18.00 & 8.11 \\
\Conf $\tau$=0.95~\cite{hu2022where2comm} & S & 5.0 & 0 & 31.56 & 18.59 & 8.48 \\
\Conf $\tau$=0.90~\cite{hu2022where2comm} & S & 9.7 & 0 & 32.21 & 19.19 & 9.18 \\
\EgoTemp $\tau$=0.98$^\dagger$ (CooperTrim-style~\cite{mukhopadhyay2026coopertrim}) & R & 3.0 & 0 & 32.87 & 20.46 & 9.23 \\
\EgoTemp $\tau$=0.95$^\dagger$ (CooperTrim-style~\cite{mukhopadhyay2026coopertrim}) & R & 6.6 & 0 & 33.97 & 21.52 & 9.99 \\
JEPA error $\tau$=0.99$^\dagger$ & R & 3.3 & 0 & 32.10 & 19.47 & 8.52 \\
\Rand (as $\tau$=0.98)$^\dagger$ & -- & 2.7 & 0 & 31.33 & 18.40 & 8.00 \\
\Rand (as $\tau$=0.95)$^\dagger$ & -- & 6.1 & 0 & 32.20 & 19.33 & 8.71 \\
GT-moving oracle$^\dagger$ & GT & 5.9 & 0 & 38.76 & 26.87 & 19.96 \\
\midrule
\Change $\tau$=0.95 + $W$ fill$^\dagger$ & S+R & 6.1 & 0 & 34.12 & 20.13 & 19.64 \\
\Native, delay-free & -- & -- & 0 & 42.50 & -- & -- \\
\bottomrule
\end{tabular}
\end{adjustbox}
\end{table*}

\section{Predictor $W$}
\label{sec:supp_train}

\paragraph{Inputs.}
For each link, $W$ receives the last four end-of-frame states of the mirror, each warped from its own ego frame into the current one; missing slots are left empty. It also receives the age of each cell's feature, a support mask, the four relative poses as $3\times4$ matrices, the current frame index and the sender type (CAV or RSU). The three scales have $96\times352$, $48\times176$ and $24\times88$ cells with 64, 128 and 256 channels.

\paragraph{Architecture.}
Each scale projects its features to 64 channels with a $1\times1$ convolution and reduces the concatenated history, ages, masks, poses and type to 64 channels with a $1\times1$ convolution, group normalization and SiLU. A trunk at the coarsest scale has three downsampling and three upsampling blocks with skip connections; each block has two $3\times3$ convolutions with group normalization and SiLU. Lateral $1\times1$ convolutions pass the trunk output from coarse to fine. At each scale, a $1\times1$ convolution outputs one 2D displacement per history slot, limited by a radial tanh to 2\,m per frame of cell age (20\,m/s), which is 2.5, 1.25 and 0.625 cells per frame at the three scales. Features are warped by bilinear sampling with zero padding, and support and age maps by nearest-neighbor sampling. Where several history slots support a cell, the prediction takes the transported feature that was captured most recently. The checkpoint has 648,539 parameters, 559,552 of them in the trunk. The protocol uses the warped features and their support only; the residual and uncertainty heads of the checkpoint are evaluated but not used.

\paragraph{Training.}
$W$ starts from a base predictor (3,259 updates) and is trained for 3,150 further updates: two passes over 6,298 queries from seven training scenes, with batch size 4. We use AdamW with weight decay $10^{-4}$ and a peak learning rate of $10^{-4}$. In this final stage, the target is the collaborator's current feature after CoSDH's compression and decoding, expressed in the receiver's current frame; the sender verifies predictions against the same kind of feature in the protocol. The loss is the mean squared error in feature space, together with a Gaussian negative log-likelihood that trains the uncertainty heads. It covers cells in which the receiver has fewer than four LiDAR points and the collaborator at least four. The feature age is drawn uniformly from 1 to 20 frames. The base predictor is trained from a random initialization toward the same target. Like the final stage, it uses no ground-truth labels and leaves the detector frozen. Its squared error is normalized per channel and weighted in each cell by one plus the frozen detector's classification confidence, over the cells that a stored feature can reach at 20\,m/s, and its input is up to four earlier messages of the collaborator.

The final stage differs from the protocol in several ways. It fills only the first history slot, with the features that the collaborator would have sent at the earlier time, encoded directly in the current frame; the protocol fills up to four slots with mirror states. It feeds the receiver's current feature and earlier features of the other collaborators as context; the protocol sets these inputs to zero, because the sender does not have the receiver's current feature and both ends must run the same computation. Its loss is applied to the transported features plus a learned residual, whereas the protocol uses the transported features alone. It supervises only the cells named above, with a squared error; the protocol verifies every candidate cell with the mean absolute difference of \cref{sec:modes}, against the sender's current feature, which is also multiplied by the supply mask (\cref{sec:supp_protocol}).

\section{Additional results}
\label{sec:supp_tables}

\subsection{Every-frame protocol, all metrics}
\label{sec:supp_everyframe}

\Cref{tab:supp_test} extends \cref{tab:main} to all four metrics and to 20\% packet loss. For each policy, the recovered share differs by at most 3.4 points across metrics.

\begin{table*}[!htbp]
\caption{\textbf{Every-frame protocol, test set, all metrics} (3,936 queries). Rec.: share of the delay loss recovered (\%), with the loss-free anchors for all rows. Paired intervals for \Change at $\tau=0.95$ against the native message, \Conf and \Rand are in \cref{tab:supp_block} and \cref{tab:supp_t1_scene}; the remaining entries are point estimates.}
\label{tab:supp_test}
\centering
\footnotesize
\setlength{\tabcolsep}{4pt}
\begin{adjustbox}{max width=\linewidth}
\begin{tabular}{@{}lrrrrrrrrr@{}}
\toprule
& & \multicolumn{2}{c}{BEV AP@0.5} & \multicolumn{2}{c}{BEV AP@0.7} & \multicolumn{2}{c}{3D AP@0.5} & \multicolumn{2}{c}{3D AP@0.7} \\
\cmidrule(lr){3-4}\cmidrule(lr){5-6}\cmidrule(lr){7-8}\cmidrule(lr){9-10}
Method & $r$ (\%) & AP & Rec. & AP & Rec. & AP & Rec. & AP & Rec. \\
\midrule
\Native, delay-free & -- & 70.27 & 100 & 46.96 & 100 & 55.93 & 100 & 23.78 & 100 \\
\Native (C-V2X) & 0 & 54.43 & 0 & 34.20 & 0 & 42.14 & 0 & 16.99 & 0 \\
+ \Rand & 5.79 & 55.72 & 8.1 & 35.13 & 7.3 & 43.25 & 8.0 & 17.53 & 8.0 \\
+ \Conf~\cite{hu2022where2comm} & 3.89 & 55.12 & 4.3 & 34.77 & 4.5 & 42.60 & 3.4 & 17.25 & 3.9 \\
+ \Change ($\tau$=0.98) & 2.46 & 65.80 & 71.8 & 43.49 & 72.8 & 52.03 & 71.7 & 21.83 & 71.3 \\
+ \Change ($\tau$=0.95) & 5.79 & 66.87 & 78.5 & 44.58 & 81.4 & 52.98 & 78.6 & 22.55 & 81.9 \\
\midrule
\multicolumn{10}{@{}l}{\textit{20\% packet loss on every message}} \\
\Native (C-V2X) & 0 & 54.42 & $-$0.1 & 34.20 & 0.0 & 42.15 & 0.1 & 16.95 & $-$0.6 \\
+ \Rand & 5.79 & 55.47 & 6.5 & 34.92 & 5.6 & 43.01 & 6.4 & 17.39 & 5.9 \\
+ \Change ($\tau$=0.95) & 5.79 & 64.33 & 62.5 & 42.42 & 64.4 & 50.81 & 62.9 & 21.38 & 64.7 \\
\bottomrule
\end{tabular}
\end{adjustbox}
\end{table*}

\subsection{Keyframe mode on DEV}
\label{sec:supp_dev}

\Cref{tab:supp_dev} gives the keyframe protocols on DEV.

\begin{table*}[!htbp]
\caption{\textbf{Keyframe mode on DEV} (2,406 queries). Paired differences with 95\% CIs; bold: CI excludes 0. Bytes are per link-frame. $^\S$Not registered.}
\label{tab:supp_dev}
\centering
\footnotesize
\setlength{\tabcolsep}{5pt}
\begin{adjustbox}{max width=\linewidth}
\begin{tabular}{@{}lccccrr@{}}
\toprule
& BEV AP@0.5 & BEV AP@0.7 & 3D AP@0.5 & 3D AP@0.7 & Down (KB) & Up (KB) \\
\midrule
Native every frame + \Change & 61.91 & 46.19 & 52.25 & 20.77 & 44.6 & 14.0 \\
\ChangeKF (CoST-style~\cite{tang2025cost}) & 60.40 & 44.35 & 50.54 & 19.40 & 6.0 & 7.1 \\
\ChangeM & 62.22 & 45.34 & 51.89 & 19.65 & 6.0 & 7.1 \\
\ours & 62.15 & 45.40 & 52.04 & 19.98 & 6.0 & 7.1 \\
\addlinespace[2pt]
\ours $-$ \ChangeKF & \textbf{+1.75} \ci{1.46, 2.02} & \textbf{+1.05} \ci{0.71, 1.38} & \textbf{+1.50} \ci{1.18, 1.82} & \textbf{+0.59} \ci{0.29, 0.91} & & \\
\ChangeM $-$ \ChangeKF & \textbf{+1.82} \ci{1.55, 2.09} & \textbf{+0.99} \ci{0.65, 1.31} & \textbf{+1.35} \ci{1.02, 1.69} & +0.25 \ci{$-$0.03, 0.56} & & \\
\ours $-$ \ChangeM & $-$0.07 \ci{$-$0.22, 0.07} & +0.06 \ci{$-$0.14, 0.28} & +0.15 \ci{$-$0.03, 0.32} & \textbf{+0.33} \ci{0.12, 0.54} & & \\
\addlinespace[2pt]
\ChangeKF $-$ every frame$^\S$ & \textbf{$-$1.51} \ci{$-$1.79, $-$1.25} & \textbf{$-$1.84} \ci{$-$2.17, $-$1.52} & \textbf{$-$1.71} \ci{$-$2.02, $-$1.39} & \textbf{$-$1.37} \ci{$-$1.68, $-$1.07} & & \\
\ChangeM $-$ every frame$^\S$ & \textbf{+0.31} \ci{0.03, 0.57} & \textbf{$-$0.85} \ci{$-$1.23, $-$0.54} & \textbf{$-$0.36} \ci{$-$0.68, $-$0.03} & \textbf{$-$1.12} \ci{$-$1.43, $-$0.77} & & \\
\ours $-$ every frame$^\S$ & +0.24 \ci{$-$0.05, 0.51} & \textbf{$-$0.79} \ci{$-$1.14, $-$0.46} & $-$0.21 \ci{$-$0.57, 0.11} & \textbf{$-$0.78} \ci{$-$1.11, $-$0.43} & & \\
\bottomrule
\end{tabular}
\end{adjustbox}
\end{table*}

\subsection{AP by object speed}
\label{sec:supp_speed}

\Cref{tab:supp_speed_test} gives the test values behind \cref{fig:gain}a,b for all four metrics, together with the predictor's own increment. The speed label requires a unique vehicle ID and an unambiguous position 0.5\,s earlier. Unmatched predictions count as false positives in every bin.

\begin{table*}[!htbp]
\caption{\textbf{Paired differences by object speed, test set} (registered). 95\% CIs; bold: CI excludes 0. GT: ground-truth objects in the bin; 1,881 of the 70,619 objects (2.7\%) have no speed label and are left out of the bins.}
\label{tab:supp_speed_test}
\centering
\footnotesize
\setlength{\tabcolsep}{4pt}
\begin{adjustbox}{max width=\linewidth}
\begin{tabular}{@{}lrlcccc@{}}
\toprule
Speed (m/s) & GT & Difference & BEV AP@0.5 & BEV AP@0.7 & 3D AP@0.5 & 3D AP@0.7 \\
\midrule
0--2 & 50,555 & \ours $-$ \ChangeKF & \textbf{+2.00} [1.70, 2.30] & \textbf{+1.64} [1.32, 1.94] & \textbf{+3.53} [3.18, 3.87] & +0.10 [$-$0.19, 0.39] \\
 & & \ChangeM $-$ \ChangeKF & \textbf{+1.90} [1.60, 2.19] & \textbf{+1.25} [0.94, 1.55] & \textbf{+2.85} [2.52, 3.19] & +0.22 [$-$0.06, 0.51] \\
 & & \ours $-$ \ChangeM & +0.10 [$-$0.06, 0.27] & \textbf{+0.39} [0.20, 0.60] & \textbf{+0.68} [0.47, 0.88] & $-$0.12 [$-$0.33, 0.07] \\
\midrule
2--5 & 5,200 & \ours $-$ \ChangeKF & \textbf{+8.47} [7.72, 9.16] & \textbf{+1.70} [1.46, 1.94] & \textbf{+4.32} [3.92, 4.74] & \textbf{+0.24} [0.14, 0.33] \\
 & & \ChangeM $-$ \ChangeKF & \textbf{+7.75} [7.03, 8.42] & \textbf{+1.41} [1.18, 1.66] & \textbf{+3.62} [3.27, 4.06] & \textbf{+0.20} [0.10, 0.29] \\
 & & \ours $-$ \ChangeM & \textbf{+0.72} [0.30, 1.09] & \textbf{+0.29} [0.12, 0.44] & \textbf{+0.70} [0.46, 0.92] & +0.04 [$-$0.02, 0.11] \\
\midrule
5--10 & 8,944 & \ours $-$ \ChangeKF & \textbf{+11.10} [10.46, 11.73] & \textbf{+2.86} [2.60, 3.14] & \textbf{+6.01} [5.62, 6.45] & \textbf{+0.62} [0.51, 0.75] \\
 & & \ChangeM $-$ \ChangeKF & \textbf{+10.19} [9.56, 10.79] & \textbf{+2.34} [2.10, 2.61] & \textbf{+5.15} [4.80, 5.57] & \textbf{+0.47} [0.37, 0.59] \\
 & & \ours $-$ \ChangeM & \textbf{+0.91} [0.62, 1.21] & \textbf{+0.51} [0.36, 0.67] & \textbf{+0.85} [0.66, 1.05] & \textbf{+0.14} [0.07, 0.22] \\
\midrule
$>$10 & 4,039 & \ours $-$ \ChangeKF & \textbf{+8.14} [7.47, 8.80] & \textbf{+2.07} [1.88, 2.30] & \textbf{+3.91} [3.58, 4.28] & \textbf{+0.45} [0.38, 0.55] \\
 & & \ChangeM $-$ \ChangeKF & \textbf{+7.48} [6.82, 8.12] & \textbf{+1.79} [1.59, 2.01] & \textbf{+3.26} [2.97, 3.63] & \textbf{+0.39} [0.31, 0.48] \\
 & & \ours $-$ \ChangeM & \textbf{+0.66} [0.40, 0.95] & \textbf{+0.28} [0.17, 0.41] & \textbf{+0.65} [0.47, 0.82] & \textbf{+0.06} [0.01, 0.12] \\
\bottomrule
\end{tabular}
\end{adjustbox}
\end{table*}

\subsection{False positives, ghosts and localization}
\label{sec:supp_fp}

\Cref{tab:supp_fp} lists the counts behind \cref{fig:gain}c,d. Detections have a score of at least 0.20 after NMS and are matched with 3D IoU. A false positive is a ghost if its center lies within 2\,m of the ground-truth center of a vehicle in any of the previous 1 to 17 frames and that vehicle is now at least 3\,m from this position; false positives for which this test is undefined are not counted as ghosts (1,816, 1,283 and 1,289 for \ChangeKF, \ChangeM and \ours). Of the 9,660 false positives that \ours avoids relative to \ChangeKF at IoU 0.5, 5,765 are ghosts. The predictor itself does not reduce ghosts: \ours makes 592 fewer false positives than \ChangeM but 105 more ghosts; its effect is on matches (319 more true positives at IoU 0.5). Recall at IoU 0.5 changes from 70.4\% to 69.9\% (0--2\,m/s), 58.9\% to 62.8\% (2--5), 60.6\% to 63.7\% (5--10) and 59.5\% to 61.0\% ($>$10) between \ChangeKF and \ours. The rows for the conditions of \cref{tab:main} were not registered.

\begin{table}[!htbp]
\caption{\textbf{False positives (FP) and ghosts on the test set} (descriptive). True positives at IoU 0.5 are 47,281, 47,238 and 47,557 for \ChangeKF, \ChangeM and \ours. Dashes: not extracted. $^\S$Not registered.}
\label{tab:supp_fp}
\centering
\footnotesize
\setlength{\tabcolsep}{3pt}
\begin{adjustbox}{max width=\linewidth}
\begin{tabular}{@{}lrrrr@{}}
\toprule
& \multicolumn{2}{c}{IoU 0.5} & \multicolumn{2}{c}{IoU 0.7} \\
\cmidrule(lr){2-3}\cmidrule(lr){4-5}
& FP & ghosts & FP & TP \\
\midrule
\Native, delay-free$^\S$ & 32,915 & 467 & -- & -- \\
\Native, 8 frames late$^\S$ & 42,846 & 9,899 & -- & -- \\
\Native every frame + \Change$^\S$ & 39,041 & 6,086 & -- & -- \\
\midrule
\ChangeKF & 39,529 & 6,429 & 58,901 & 27,909 \\
\ChangeM & 30,461 & 559 & 50,104 & 27,595 \\
\ours & 29,869 & 664 & 49,690 & 27,736 \\
\bottomrule
\end{tabular}
\end{adjustbox}
\end{table}

We also follow each ground-truth object through its best-matching box and say that a method localizes it if the 3D IoU is at least 0.5. Among the 14,144 objects at 2--10\,m/s, the late native message localizes 4,657 and \ours 8,961; 1,982 go from below 0.5 to at least 0.7 and 42 the other way (not registered). From \ChangeKF to \ours, 180 go from below 0.5 to at least 0.7 and 47 the other way. From \ChangeM to \ours, 671 of these objects move up an IoU bin and 489 move down. Below 2\,m/s the two directions against the late message are level, at 530 and 510, and from \ChangeKF to \ours 3,808 objects move up an IoU bin and 4,214 down, so the AP gain of near-static objects does not come from better boxes; fewer false positives are the likely source.

\subsection{Where the bytes go}
\label{sec:supp_bytes}

\Cref{tab:supp_comp} splits the increment downlink of \ChangeM and \ours on the test set. \ours sends 188,559 fewer feature cells (15.1\%), and the feature bytes that this saves equal the extra bytes of mode map to within 157\,B. In exchange, 2.0 million cells are updated by transport, 10.6 for each feature cell given up. DEV shows the same pattern (107,650 fewer feature cells, 1.12 million transported cells).

\begin{table}[!htbp]
\caption{\textbf{Composition of the increment downlink}, test set, 9,680 evaluated link-frames. Every sent cell in these runs costs 12 bytes.}
\label{tab:supp_comp}
\centering
\footnotesize
\setlength{\tabcolsep}{3pt}
\begin{adjustbox}{max width=\linewidth}
\begin{tabular}{@{}lrr@{}}
\toprule
& \ChangeM & \ours \\
\midrule
Increment downlink (B) & 18,479,839 & 18,479,996 \\
\quad \Send features and indices & 14,974,968 & 12,712,260 \\
\quad mode map & 1,603,513 & 3,866,378 \\
\quad headers & 1,901,358 & 1,901,358 \\
\midrule
\Send cells & 1,247,914 & 1,059,355 \\
\Pred cells & 0 & 2,003,989 \\
\Clear cells & 1,513,765 & 2,151,607 \\
\bottomrule
\end{tabular}
\end{adjustbox}
\end{table}

\subsection{Latency}
\label{sec:supp_latency}

\Cref{tab:supp_latency} lists the component times on an otherwise idle NVIDIA L40S. The selection step of our reference implementation re-encodes the whole mode map, including a zlib pass, after each candidate cell to check the budget (125 encodings per frame in this workload). Tracking the encoded size incrementally would remove most of this cost; we have not implemented it.

\begin{table}[!htbp]
\caption{\textbf{Median latency} (batch 1, 500 runs after warm-up, NVIDIA L40S). The rows are separate micro-benchmarks with different scopes and should not be added.}
\label{tab:supp_latency}
\centering
\footnotesize
\setlength{\tabcolsep}{3pt}
\begin{adjustbox}{max width=\linewidth}
\begin{tabular}{@{}lr@{}}
\toprule
Component & ms \\
\midrule
\multicolumn{2}{@{}l}{\textit{One query (receiver and two collaborators)}} \\
CoSDH encoding & 17.9 \\
Fusion and detection head (no NMS) & 1.9 \\
\midrule
\multicolumn{2}{@{}l}{\textit{Per link}} \\
Predictor $W$ with warp & 8.8 \\
Sender: verification and cell selection & 125.6 \\
Sender: mode-map coding and packing {\scriptsize(included in the row above)} & 2.1 \\
Sender: mirror update & 2.3 \\
Receiver: unpacking and state update & 2.6 \\
\bottomrule
\end{tabular}
\end{adjustbox}
\end{table}

At 1.6\,Mbps, sending the increments into a receiver takes 23.5\,ms on average and at most 32\,ms (4.7 and 6.4\,KB on the test set).

\subsection{Qualitative examples}
\label{sec:supp_qual}

\begin{figure*}[!htbp]
\centering
\includegraphics[width=0.86\textwidth]{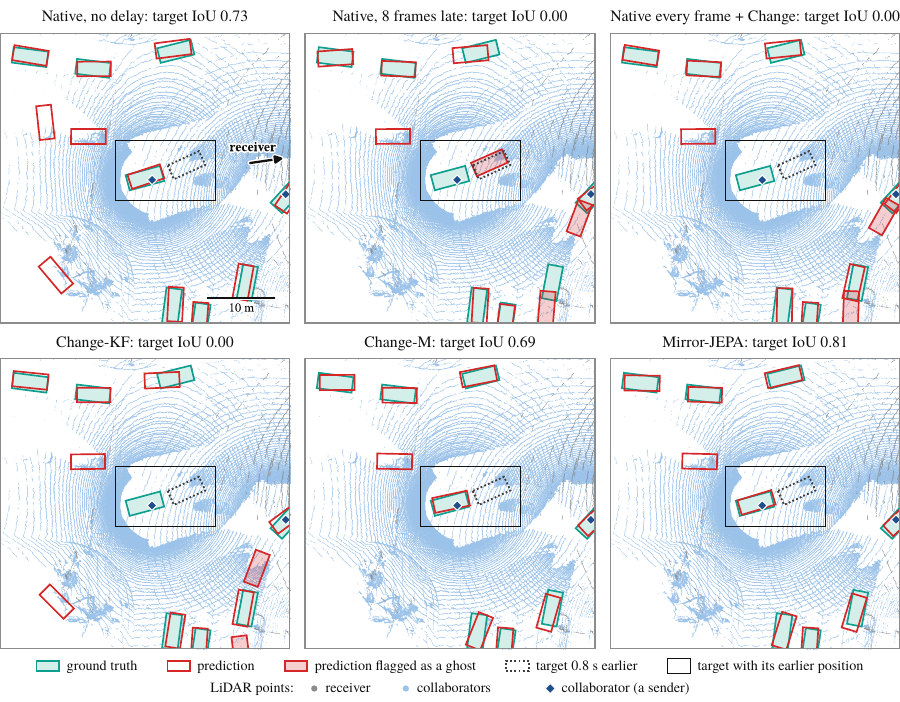}
\caption{\textbf{The other candidate query.} Titles give the best 3D IoU of a prediction with the target, the vehicle in the black frame; 0.00 means that no prediction overlaps it. The target moves at 8.2\,m/s and covers 6.4\,m in 0.8\,s. It is itself a collaborator, CAV 3 (diamond on the target); the RSU lies outside the view and is not marked. Of the 263 LiDAR points inside its box, 255 come from the LiDAR of CAV 2 and 8 from the receiver. Scene (18 objects), matched\,/\,unmatched predictions in panel order: 13\,/\,10, 11\,/\,11, 13\,/\,9, 12\,/\,13, 13\,/\,8 and 14\,/\,7. Points are the LiDAR returns of the current frame in every panel, gray from the receiver and blue from the collaborators.}
\label{fig:supp_qual2}
\end{figure*}

\paragraph{Selection rule.}
A ground-truth object qualifies as an example if (a) its best 3D IoU under the native message that is eight frames late is below 0.5 and an unmatched prediction lies where the object stood 0.8\,s earlier, (b) its best 3D IoU under \ChangeKF is below 0.5, (c) its best 3D IoU under \ours is at least 0.7, and (d) it moves at 2--10\,m/s. Of the 14,144 test objects in this speed range, 88 meet all four conditions. We had fixed two candidate queries earlier, among the 32 objects with a best IoU below 0.5 under \ChangeKF, between 0.5 and 0.7 under \ChangeM and at least 0.7 under \ours; both meet the four conditions. The target of the first is itself a cooperating vehicle, whose pose a deployed system could broadcast directly, so we added a fifth condition after the selection, that the target is not a cooperating vehicle; 69 of the 88 objects remain. \Cref{fig:method} shows the second candidate; we report the first here as well (\cref{fig:supp_qual2}).

\paragraph{Query of \cref{fig:method}.}
The scene has 22 ground-truth objects and three collaborators, two CAVs and one RSU; the RSU lies outside the view and is not marked. None of the four agents lies inside the target's box. The target moves at 6.6\,m/s. The keyframes that \ours holds in this frame were captured 14 frames earlier. Over the three links it sends 354 cells, transports 764 and clears 766, in 6.4\,KB of increments. Without delay the best IoU with the target is 0.60 (20 matched and 5 unmatched predictions in the scene), below the 0.74 of \ours. One of the four additional matches in \cref{fig:method}c is a cooperating vehicle (diamond). \Cref{fig:supp_qual2} shows the other candidate under six conditions, as a view from above in a 42.7\,m window centered on the target, drawn like \cref{fig:method}. The black frame marks the target and its box 0.8\,s earlier (dotted). The receiver lies outside every window, and an arrow in the first panel points to it. The comparisons with the three conditions of \cref{tab:main} in the top rows were not registered.

\begin{figure*}[!htbp]
\centering
\includegraphics[width=0.8\textwidth]{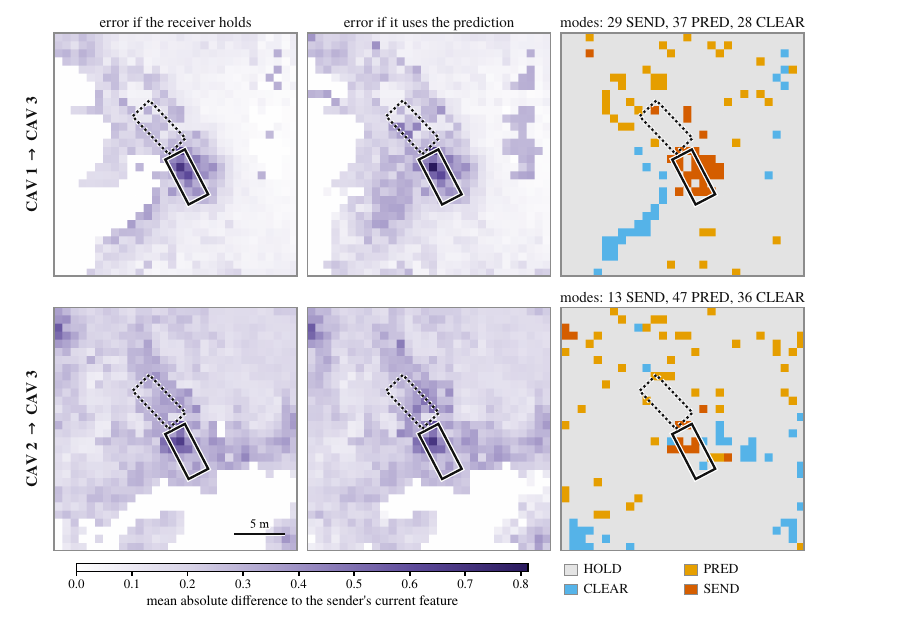}
\caption{\textbf{The cells behind \cref{fig:method}} for the two vehicle links, in a 24\,m window around the target (solid: now; dotted: 0.8\,s earlier), at the finest scale. Left: mean absolute difference between the sender's current feature and the feature that the receiver holds. Middle: the same for the predictor's output. Right: the modes that the sender chose. The holding error is largest on the moving vehicle, and the prediction does not lower it there, so the sender transmits features on the vehicle and marks predictions and clears around it. In this window the three senders transmit 44 feature cells and mark 85 predictions and 67 clears; the RSU link accounts for 2, 1 and 3 of them.}
\label{fig:supp_cells}
\end{figure*}

\paragraph{Cells of this frame.}
Over the full map and the three scales, the link from CAV 1 carries 113 \Send, 265 \Pred and 292 \Clear cells, the link from CAV 2 carries 99, 442 and 344, and the link from the RSU 142, 57 and 130. All \Send cells are at the finest scale. \Cref{fig:supp_cells} shows the window around the target for the two vehicle links. Inside the target's current box, each of the two vehicle links has 17 cells at the finest scale. Together they are 16 \Send, 17 \Hold and 1 \Clear, and none is \Pred: the prediction error there averages 0.431 and 0.387 on the two links and is not below the holding error, which averages 0.424 and 0.387, against 0.092 and 0.172 in the rest of the window outside the target's current and earlier boxes.

\section{Registration and amendments}
\label{sec:supp_hygiene}

All criteria, thresholds, budgets and decision rules of the keyframe comparisons were registered with timestamps. Each registration lists the earlier results we had seen. We report all of them here.

\paragraph{Test set.}
We registered the test criteria of \cref{tab:supp_reg_test} at 2026-11-07 03:59:29Z, prior to  prediction or results of these runs; the record discloses that we had seen all results on DEV and on the diagnostic subset. The formal evaluation ran once, on 2026-11-09, on the same 3,936 queries as \cref{tab:main}. Between registration and the run, and before any prediction, we made five implementation fixes to import, deployment and split-adapter bindings and, after a host crash, registered an amendment that moved the official replay to a second cluster. Model weights, the budget rule, mode selection and coding, evaluation, bootstrap and criteria did not change. The hash of the evaluation script had not been recorded on its own at registration; the script reproduces the twelve registered DEV APs exactly. Each protocol received the one registered byte correction (\cref{sec:supp_protocol}). The three protocols of a sequence always run on the same GPU; across the GPU types we used, the total bytes of a 50-frame DEV replay differ by at most 0.0025\%, and a repeated run on the same GPU is identical bit for bit. The amendment kept the original host (NVIDIA RTX A6000) as a replication, which needed one further fix to a module path and was evaluated once, after the official run. Its twelve APs lie within 0.04 of the official ones and the twelve paired differences among the three protocols within 0.07; every registered criterion has the same outcome, each CI excludes zero in the same cases, and the mirror check again gives 0. The initial budget of the official run is one byte below the registered input (1,908\,B against 1,909\,B), because the native bytes measured on the second cluster are 0.01\% lower. After the byte correction, the budget of \ChangeKF is 2,138\,B in the official run and 2,139\,B in the replication, which admits one more cell per increment; \ChangeKF therefore sends 0.07\% fewer increment bytes than the other two protocols in the official run and 0.5\% more in the replication.

\begin{table*}[!htbp]
\caption{\textbf{Registered criteria on the test set} and their outcomes. Moving bins: 2--5, 5--10 and $>$10\,m/s.}
\label{tab:supp_reg_test}
\centering
\footnotesize
\setlength{\tabcolsep}{4pt}
\begin{adjustbox}{max width=\linewidth}
\begin{tabular}{@{}p{2.5cm}p{7.6cm}p{6.4cm}@{}}
\toprule
Criterion & Rule (\ours $=$ M, \ChangeKF $=$ K, \ChangeM $=$ C) & Test result \\
\midrule
3D (primary) & M $-$ K on 3D AP@0.7 positive with CI lower bound above 0 & pass: +0.54 [0.30, 0.81] \\
BEV & primary: M $-$ K on BEV AP@0.7 positive with CI above 0; secondary: BEV AP@0.5 and 3D AP@0.5 positive & pass: +2.21 [1.93, 2.48]; +2.98, +3.97 \\
BEV speed bins & primary: M $-$ K on BEV AP@0.7 positive in 5--10 and $>$10\,m/s, at least one CI above 0; secondary: positive in 2--5\,m/s & pass: +2.86 [2.60, 3.14], +2.07 [1.88, 2.30]; +1.70 \\
3D speed bins & primary: M $-$ K on 3D AP@0.7 positive in 5--10 and $>$10\,m/s, at least one CI above 0; secondary: M $-$ C positive in all three moving bins & pass: +0.62 [0.51, 0.75], +0.45 [0.38, 0.55]; +0.04, +0.14, +0.06 \\
Predictor (descriptive) & M $-$ C, all four metrics & +0.30 [0.17, 0.45], +0.54 [0.38, 0.71], +0.73 [0.57, 0.90], +0.03 [$-$0.14, 0.19] \\
\bottomrule
\end{tabular}
\end{adjustbox}
\end{table*}

\paragraph{Mirror check.}
For \ChangeM and \ours, the replay compares the sender's mirror with the receiver state in every frame. On the test set this covers 60 completed walks with 38,810 replayed frames and 95,446 active link-frames per protocol, and the maximum absolute difference is 0. On DEV the check covers all 2,406 queries and 6,193 links, with 23,484 sequential frame visits in the final pass, and the maximum absolute difference is 0.

\paragraph{DEV.}
\Cref{tab:supp_reg} lists the four DEV criteria. All were registered before the DEV AP files were generated (2026-11-06 22:41:48Z), and their records disclose that we had seen the results on the diagnostic subset. Two parts fail on DEV: the predictor's increment under 3D AP@0.7 falls short of the registered 0.7 points, and it is not positive above 10\,m/s. The second part passes on the test set.

\begin{table*}[!htbp]
\caption{\textbf{Registered DEV criteria} and their outcomes. Moving bins: 2--5, 5--10 and $>$10\,m/s.}
\label{tab:supp_reg}
\centering
\footnotesize
\setlength{\tabcolsep}{4pt}
\begin{adjustbox}{max width=\linewidth}
\begin{tabular}{@{}p{2.6cm}p{2.3cm}p{6.6cm}p{4.6cm}@{}}
\toprule
Criterion & Registered & Rule (\ours $=$ M, \ChangeKF $=$ K, \ChangeM $=$ C) & DEV result \\
\midrule
Predictor & 2026-11-05 13:40:25Z & M $-$ C on 3D AP@0.7 at least 0.7 with CI above 0 & fail: +0.33 [0.12, 0.54] \\
3D speed bins & proposed 2026-11-06 02:06Z & primary: M $-$ K on 3D AP@0.7 positive in 5--10 and $>$10\,m/s, at least one CI above 0; secondary: M $-$ C positive in all three moving bins & primary pass: +0.45 [0.31, 0.61], +0.30 [0.23, 0.40]; secondary fail: $-$0.01 in $>$10\,m/s \\
BEV & 2026-11-06 16:30:56Z & primary: M $-$ K on BEV AP@0.7 positive with CI above 0; secondary: BEV AP@0.5 and 3D AP@0.5 positive & pass: +1.05 [0.71, 1.38]; +1.75, +1.50 \\
BEV speed bins & 2026-11-06 17:34:47Z & primary: M $-$ K on BEV AP@0.7 positive in 5--10 and $>$10\,m/s, at least one CI above 0; secondary: positive in 2--5\,m/s & pass: +2.26 [1.85, 2.62], +1.14 [0.93, 1.40]; +1.86 [1.48, 2.25] \\
\bottomrule
\end{tabular}
\end{adjustbox}
\end{table*}

\begin{table*}[!htbp]
\caption{\textbf{Keyframe protocols within each test scene.} Paired differences with 95\% CIs from resampling the queries of the scene; bold: CI excludes 0. Every frame: the native message streamed every frame with the \Change delta.}
\label{tab:supp_scene}
\centering
\footnotesize
\setlength{\tabcolsep}{5pt}
\begin{adjustbox}{max width=\linewidth}
\begin{tabular}{@{}llcccc@{}}
\toprule
Scene & Difference & BEV AP@0.5 & BEV AP@0.7 & 3D AP@0.5 & 3D AP@0.7 \\
\midrule
2 (2,643 queries) & \ours $-$ \ChangeKF & \textbf{+2.66} \ci{2.33, 3.00} & \textbf{+1.91} \ci{1.61, 2.24} & \textbf{+4.06} \ci{3.69, 4.40} & $-$0.06 \ci{$-$0.30, 0.22} \\
 & \ChangeM $-$ \ChangeKF & \textbf{+2.32} \ci{2.03, 2.63} & \textbf{+1.17} \ci{0.87, 1.50} & \textbf{+3.17} \ci{2.81, 3.49} & $-$0.05 \ci{$-$0.27, 0.22} \\
 & \ours $-$ \ChangeM & \textbf{+0.35} \ci{0.16, 0.51} & \textbf{+0.74} \ci{0.54, 0.94} & \textbf{+0.89} \ci{0.68, 1.10} & $-$0.01 \ci{$-$0.20, 0.16} \\
 & \ours $-$ every frame & \textbf{+1.84} \ci{1.53, 2.16} & \textbf{+0.55} \ci{0.23, 0.90} & \textbf{+3.27} \ci{2.89, 3.65} & \textbf{$-$0.95} \ci{$-$1.20, $-$0.70} \\
\midrule
8 (1,293 queries) & \ours $-$ \ChangeKF & \textbf{+3.66} \ci{3.22, 4.07} & \textbf{+2.90} \ci{2.39, 3.38} & \textbf{+3.48} \ci{3.01, 3.93} & \textbf{+2.45} \ci{1.84, 3.02} \\
 & \ChangeM $-$ \ChangeKF & \textbf{+3.41} \ci{2.96, 3.82} & \textbf{+2.76} \ci{2.26, 3.24} & \textbf{+3.12} \ci{2.64, 3.53} & \textbf{+2.22} \ci{1.64, 2.83} \\
 & \ours $-$ \ChangeM & \textbf{+0.25} \ci{0.01, 0.48} & +0.14 \ci{$-$0.15, 0.42} & \textbf{+0.36} \ci{0.11, 0.60} & +0.22 \ci{$-$0.24, 0.64} \\
 & \ours $-$ every frame & \textbf{+2.35} \ci{1.90, 2.75} & \textbf{+0.94} \ci{0.46, 1.47} & \textbf{+2.02} \ci{1.52, 2.50} & +0.41 \ci{$-$0.19, 1.03} \\
\bottomrule
\end{tabular}
\end{adjustbox}
\end{table*}

\begin{table*}[!htbp]
\caption{\textbf{\ours $-$ \ChangeKF by object speed within each test scene.} 95\% CIs; bold: CI excludes 0. Unmatched predictions count as false positives in every bin, as in \cref{tab:supp_speed_test}.}
\label{tab:supp_scene_speed}
\centering
\footnotesize
\setlength{\tabcolsep}{4pt}
\begin{adjustbox}{max width=\linewidth}
\begin{tabular}{@{}lllcccc@{}}
\toprule
Scene & Speed (m/s) & Difference & BEV AP@0.5 & BEV AP@0.7 & 3D AP@0.5 & 3D AP@0.7 \\
\midrule
2 & 0--2 & \ours $-$ \ChangeKF & \textbf{+1.64} \ci{1.26, 2.01} & \textbf{+1.23} \ci{0.88, 1.59} & \textbf{+3.81} \ci{3.39, 4.19} & \textbf{$-$0.62} \ci{$-$0.90, $-$0.33} \\
 & 2--5 & \ours $-$ \ChangeKF & \textbf{+7.23} \ci{6.48, 7.98} & \textbf{+1.33} \ci{1.10, 1.58} & \textbf{+3.22} \ci{2.85, 3.64} & \textbf{+0.21} \ci{0.12, 0.31} \\
 & 5--10 & \ours $-$ \ChangeKF & \textbf{+9.87} \ci{9.15, 10.57} & \textbf{+2.44} \ci{2.15, 2.72} & \textbf{+4.49} \ci{4.16, 4.93} & \textbf{+0.45} \ci{0.33, 0.59} \\
 & $>$10 & \ours $-$ \ChangeKF & \textbf{+8.06} \ci{7.33, 8.82} & \textbf{+2.05} \ci{1.83, 2.29} & \textbf{+3.70} \ci{3.35, 4.11} & \textbf{+0.46} \ci{0.37, 0.58} \\
\midrule
8 & 0--2 & \ours $-$ \ChangeKF & \textbf{+2.73} \ci{2.22, 3.21} & \textbf{+2.56} \ci{1.98, 3.15} & \textbf{+2.63} \ci{2.10, 3.13} & \textbf{+2.55} \ci{1.80, 3.25} \\
 & 2--5 & \ours $-$ \ChangeKF & \textbf{+12.71} \ci{11.06, 14.00} & \textbf{+4.07} \ci{2.92, 5.17} & \textbf{+9.75} \ci{8.34, 11.09} & \textbf{+0.45} \ci{0.05, 0.84} \\
 & 5--10 & \ours $-$ \ChangeKF & \textbf{+13.85} \ci{12.61, 14.88} & \textbf{+6.05} \ci{5.21, 6.87} & \textbf{+11.08} \ci{9.94, 12.09} & \textbf{+1.75} \ci{1.42, 2.23} \\
 & $>$10 & \ours $-$ \ChangeKF & \textbf{+8.85} \ci{7.37, 10.61} & \textbf{+2.97} \ci{2.29, 3.79} & \textbf{+6.32} \ci{5.08, 7.76} & \textbf{+0.47} \ci{0.29, 0.66} \\
\bottomrule
\end{tabular}
\end{adjustbox}
\end{table*}

\paragraph{Comparisons that were not registered.}
The comparisons with the conditions of \cref{tab:main} were not registered: the differences between \ours and the every-frame stream in \cref{tab:ladder} and \cref{tab:supp_dev}, the light gray rows of \cref{fig:gain}c, the rows marked in \cref{tab:supp_fp}, and the localization counts against the late native message. Both runs use the same 3,936 test queries. Their evaluations differ only in the floating-point implementation of the box-corner conversion, and re-evaluating the delay-free, delayed and \Change conditions of \cref{tab:main} with the keyframe implementation reproduces all twelve AP values exactly. The ghost and localization statistics are descriptive. The checks of \cref{sec:supp_robust} were not registered either. The two runs also build their messages differently: the conditions of \cref{tab:main} encode the delayed background with the request map of the current frame (\cref{sec:supp_protocol}), whereas the keyframe protocols are strictly causal.

\paragraph{Other registered outcomes.}
On the diagnostic subset (128 queries from training scenes 10 to 12), the registered criterion for the predictor's increment required a CI above zero under 3D AP@0.7; the result, $+1.12$ with 95\% CI $[-0.01,2.28]$, does not meet it, and \ours $-$ \ChangeKF was $-0.60$ ($[-1.95,0.85]$) there, where DEV shows $+0.59$ and the test set $+0.54$. A gate that restricts \Pred and \Clear to cells that the sender observes and the receiver requests failed its registered criterion on a fixed 64-query part of that subset ($-1.63$, $[-3.63,0.45]$); the method does not use it. A registered comparison of predictors in the every-frame protocol was stopped before any detection result. Mode-map coding was amended twice, each time before the AP of the corresponding run existed; \Cref{sec:supp_protocol} gives the final coding. We also ran a predictor trained with the detection loss, but not as a matched control: its acceptance test used a different distance, its candidate set was 6.6 times larger, and its checkpoint was selected on the diagnostic subset. We draw no conclusion from it.

\subsection{Robustness checks (not registered)}
\label{sec:supp_robust}

The analyses of this section were run after the test results were known. They use the test set only, and none of them was registered. CIs are 95\% paired bootstrap intervals with 1,000 resamples.

\begin{table}[!htbp]
\caption{\textbf{The two test scenes.} Distances are horizontal, from the receiver to object centers.}
\label{tab:supp_scene_facts}
\centering
\footnotesize
\setlength{\tabcolsep}{3pt}
\begin{adjustbox}{max width=\linewidth}
\begin{tabular}{@{}lrr@{}}
\toprule
& Scene 2 & Scene 8 \\
\midrule
Queries & 2,643 & 1,293 \\
Ground-truth objects & 47,287 & 23,332 \\
Objects per query (mean) & 17.9 & 18.0 \\
Queries with no collaborators & 66 & 29 \\
Queries with one collaborator & 32 & 57 \\
Queries with two collaborators & 1,455 & 210 \\
Queries with three collaborators & 1,090 & 997 \\
Object distance, median (m) & 36.2 & 34.2 \\
Object distance, 90th percentile (m) & 74.9 & 64.7 \\
\bottomrule
\end{tabular}
\end{adjustbox}
\end{table}

\begin{table*}[!htbp]
\caption{\textbf{AP with the boxes of cooperating vehicles ignored}, test set. Paired differences with 95\% CIs; bold: CI excludes 0. Every frame: the native message streamed every frame with the \Change delta.}
\label{tab:supp_cav_ap}
\centering
\footnotesize
\setlength{\tabcolsep}{5pt}
\begin{adjustbox}{max width=\linewidth}
\begin{tabular}{@{}llcccc@{}}
\toprule
& & BEV AP@0.5 & BEV AP@0.7 & 3D AP@0.5 & 3D AP@0.7 \\
\midrule
\multicolumn{2}{@{}l}{\Native, delay-free} & 67.46 & 43.89 & 53.90 & 22.67 \\
\multicolumn{2}{@{}l}{\Native, 8 frames late} & 51.18 & 31.52 & 39.85 & 16.29 \\
\multicolumn{2}{@{}l}{\Native every frame + \Change} & 63.90 & 41.66 & 50.97 & 21.55 \\
\multicolumn{2}{@{}l}{\ChangeKF} & 62.97 & 40.22 & 50.04 & 20.46 \\
\multicolumn{2}{@{}l}{\ChangeM} & 65.62 & 41.70 & 53.25 & 20.93 \\
\multicolumn{2}{@{}l}{\ours} & 65.96 & 42.16 & 53.99 & 20.91 \\
\addlinespace[2pt]
\multicolumn{2}{@{}l}{\ours $-$ \ChangeKF} & \textbf{+2.99} \ci{2.71, 3.26} & \textbf{+1.94} \ci{1.64, 2.21} & \textbf{+3.95} \ci{3.62, 4.26} & \textbf{+0.45} \ci{0.20, 0.69} \\
\multicolumn{2}{@{}l}{\ChangeM $-$ \ChangeKF} & \textbf{+2.65} \ci{2.37, 2.92} & \textbf{+1.48} \ci{1.20, 1.75} & \textbf{+3.21} \ci{2.90, 3.51} & \textbf{+0.47} \ci{0.23, 0.72} \\
\multicolumn{2}{@{}l}{\ours $-$ \ChangeM} & \textbf{+0.34} \ci{0.19, 0.49} & \textbf{+0.46} \ci{0.29, 0.64} & \textbf{+0.74} \ci{0.56, 0.92} & $-$0.02 \ci{$-$0.21, 0.14} \\
\multicolumn{2}{@{}l}{\ours $-$ every frame} & \textbf{+2.05} \ci{1.78, 2.33} & \textbf{+0.49} \ci{0.20, 0.78} & \textbf{+3.02} \ci{2.67, 3.37} & \textbf{$-$0.64} \ci{$-$0.91, $-$0.36} \\
\bottomrule
\end{tabular}
\end{adjustbox}
\end{table*}

\begin{table*}[!htbp]
\caption{\textbf{Per-query and block-bootstrap intervals}, test set, all objects. Each comparison has two rows: the difference with its 95\% CI from resampling queries, then the CI from resampling blocks of 50 consecutive queries. Top: keyframe protocols (every frame: the native message streamed every frame with the \Change delta). Bottom: conditions of \cref{tab:main}, with \Change at $\tau=0.95$.}
\label{tab:supp_block}
\centering
\footnotesize
\setlength{\tabcolsep}{6pt}
\begin{adjustbox}{max width=\linewidth}
\begin{tabular}{@{}llcccc@{}}
\toprule
Difference & CI from & BEV AP@0.5 & BEV AP@0.7 & 3D AP@0.5 & 3D AP@0.7 \\
\midrule
\ours $-$ \ChangeKF & queries & +2.98 \ci{2.72, 3.24} & +2.21 \ci{1.93, 2.48} & +3.97 \ci{3.66, 4.25} & +0.54 \ci{0.30, 0.81} \\
 & blocks & \ci{2.12, 3.91} & \ci{1.42, 3.01} & \ci{2.80, 5.16} & \ci{$-$0.08, 1.27} \\
\addlinespace[2pt]
\ChangeM $-$ \ChangeKF & queries & +2.68 \ci{2.40, 2.94} & +1.67 \ci{1.39, 1.94} & +3.24 \ci{2.95, 3.51} & +0.51 \ci{0.27, 0.76} \\
 & blocks & \ci{1.82, 3.56} & \ci{0.93, 2.49} & \ci{2.18, 4.24} & \ci{$-$0.11, 1.22} \\
\addlinespace[2pt]
\ours $-$ \ChangeM & queries & +0.30 \ci{0.17, 0.45} & +0.54 \ci{0.38, 0.71} & +0.73 \ci{0.57, 0.90} & +0.03 \ci{$-$0.14, 0.19} \\
 & blocks & \ci{0.11, 0.51} & \ci{0.22, 0.83} & \ci{0.44, 1.06} & \ci{$-$0.24, 0.34} \\
\addlinespace[2pt]
\ours $-$ every frame & queries & +2.01 \ci{1.76, 2.27} & +0.70 \ci{0.44, 0.99} & +3.00 \ci{2.68, 3.31} & $-$0.65 \ci{$-$0.91, $-$0.38} \\
 & blocks & \ci{1.24, 2.84} & \ci{$-$0.03, 1.36} & \ci{1.72, 4.31} & \ci{$-$1.23, $-$0.04} \\
\midrule
\Change $-$ \Native (C-V2X) & queries & +12.44 \ci{12.07, 12.78} & +10.38 \ci{10.05, 10.70} & +10.84 \ci{10.52, 11.18} & +5.57 \ci{5.35, 5.80} \\
 & blocks & \ci{10.77, 14.24} & \ci{9.02, 11.78} & \ci{9.30, 12.46} & \ci{4.67, 6.53} \\
\addlinespace[2pt]
\Change $-$ \Conf & queries & +11.75 \ci{11.41, 12.11} & +9.81 \ci{9.50, 10.13} & +10.38 \ci{10.06, 10.72} & +5.31 \ci{5.08, 5.54} \\
 & blocks & \ci{10.05, 13.50} & \ci{8.44, 11.18} & \ci{8.84, 11.93} & \ci{4.43, 6.21} \\
\addlinespace[2pt]
\Change $-$ \Rand & queries & +11.15 \ci{10.80, 11.46} & +9.45 \ci{9.15, 9.74} & +9.73 \ci{9.43, 10.05} & +5.03 \ci{4.80, 5.25} \\
 & blocks & \ci{9.62, 12.76} & \ci{8.24, 10.73} & \ci{8.30, 11.21} & \ci{4.20, 5.89} \\
\bottomrule
\end{tabular}
\end{adjustbox}
\end{table*}

\newcommand{\TabEveryFrameByScene}[1]{%
#1{\textbf{Every-frame protocol within each test scene.} AP of the conditions of \cref{tab:main}, paired differences with 95\% CIs from resampling the queries of the scene (bold: CI excludes 0), and the share of the delay loss that \Change recovers.}
\label{tab:supp_t1_scene}
\centering
\footnotesize
\setlength{\tabcolsep}{5pt}
\begin{adjustbox}{max width=\linewidth}
\begin{tabular}{@{}llcccc@{}}
\toprule
Scene & & BEV AP@0.5 & BEV AP@0.7 & 3D AP@0.5 & 3D AP@0.7 \\
\midrule
2 & \Native, delay-free & 68.92 & 42.77 & 51.12 & 19.57 \\
 & \Native (C-V2X) & 53.97 & 31.10 & 38.67 & 14.05 \\
 & + \Rand & 55.18 & 31.96 & 39.68 & 14.40 \\
 & + \Conf & 54.53 & 31.62 & 39.01 & 14.26 \\
 & + \Change ($\tau$=0.95) & 65.46 & 40.64 & 48.35 & 18.65 \\
\addlinespace[2pt]
 & \Change $-$ \Native (C-V2X) & \textbf{+11.48} \ci{11.08, 11.90} & \textbf{+9.54} \ci{9.20, 9.89} & \textbf{+9.67} \ci{9.30, 10.04} & \textbf{+4.60} \ci{4.39, 4.85} \\
 & \Change $-$ \Conf & \textbf{+10.93} \ci{10.54, 11.35} & \textbf{+9.03} \ci{8.69, 9.37} & \textbf{+9.34} \ci{8.98, 9.70} & \textbf{+4.39} \ci{4.16, 4.63} \\
 & \Change $-$ \Rand & \textbf{+10.27} \ci{9.90, 10.65} & \textbf{+8.68} \ci{8.35, 9.00} & \textbf{+8.66} \ci{8.33, 9.00} & \textbf{+4.25} \ci{4.04, 4.48} \\
 & delay-free $-$ \Native (C-V2X) & \textbf{+14.94} \ci{14.44, 15.41} & \textbf{+11.67} \ci{11.25, 12.11} & \textbf{+12.45} \ci{11.98, 12.89} & \textbf{+5.52} \ci{5.22, 5.82} \\
 & Loss recovered by \Change (\%) & 76.9 & 81.8 & 77.7 & 83.4 \\
\midrule
8 & \Native, delay-free & 73.21 & 57.53 & 67.37 & 35.19 \\
 & \Native (C-V2X) & 55.59 & 42.30 & 50.52 & 25.12 \\
 & + \Rand & 56.97 & 43.49 & 51.88 & 26.17 \\
 & + \Conf & 56.52 & 42.96 & 51.22 & 25.50 \\
 & + \Change ($\tau$=0.95) & 69.99 & 54.79 & 64.21 & 33.06 \\
\addlinespace[2pt]
 & \Change $-$ \Native (C-V2X) & \textbf{+14.40} \ci{13.73, 15.08} & \textbf{+12.49} \ci{11.79, 13.14} & \textbf{+13.70} \ci{13.09, 14.33} & \textbf{+7.94} \ci{7.41, 8.50} \\
 & \Change $-$ \Conf & \textbf{+13.48} \ci{12.82, 14.13} & \textbf{+11.83} \ci{11.17, 12.46} & \textbf{+13.00} \ci{12.37, 13.62} & \textbf{+7.56} \ci{7.04, 8.11} \\
 & \Change $-$ \Rand & \textbf{+13.02} \ci{12.35, 13.66} & \textbf{+11.29} \ci{10.68, 11.91} & \textbf{+12.33} \ci{11.72, 12.94} & \textbf{+6.89} \ci{6.37, 7.45} \\
 & delay-free $-$ \Native (C-V2X) & \textbf{+17.62} \ci{16.80, 18.45} & \textbf{+15.23} \ci{14.40, 15.97} & \textbf{+16.86} \ci{16.06, 17.59} & \textbf{+10.07} \ci{9.33, 10.78} \\
 & Loss recovered by \Change (\%) & 81.8 & 82.0 & 81.3 & 78.9 \\
\bottomrule
\end{tabular}
\end{adjustbox}
}
\begin{table*}[!htbp]
\TabEveryFrameByScene{\caption}
\end{table*}

\paragraph{By scene.}
\Cref{tab:supp_scene} repeats the comparisons of \cref{tab:ladder} within each of the two test scenes, with queries resampled within the scene, and \cref{tab:supp_scene_speed} splits the difference between \ours and \ChangeKF by object speed. \Cref{tab:supp_scene_facts} describes the two scenes.

\paragraph{Cooperating vehicles in the ground truth.}
The ground truth contains the boxes of the cooperating vehicles: 5,355 boxes of other CAVs and 2,773 of the receiver itself among the 70,619 test objects (11.5\%; 9.7\% of the objects at 2--10\,m/s). A box counts as a cooperating vehicle when an agent's pose lies inside it in the bird's-eye view. Without these boxes, 12,772 objects remain at 2--10\,m/s. The late full message localizes 4,320 of them with a best 3D IoU of at least 0.5, and \ours 8,171; 1,758 go from below 0.5 to at least 0.7 and 38 the other way, and from \ChangeKF to \ours the two directions are 137 and 43. Of \ChangeKF's 6,429 ghosts, 6,029 lie where a vehicle that is not cooperating stood, against 635 of \ours's 664. \Cref{tab:supp_cav_ap} gives AP with these boxes ignored: predictions are matched against all objects first, and then these boxes and the predictions matched to them are removed. With them ignored, \Change recovers 78.2--82.5\% of the delay loss across the four metrics, against 78.5--81.9\% with all objects.

\paragraph{Block bootstrap.}
Consecutive queries are not independent. Each receiver's queries in a scene are sorted by time and cut into blocks of 50 consecutive queries (5\,s); the 75 blocks are resampled with replacement 1,000 times. \Cref{tab:supp_block} compares the resulting intervals with those from resampling single queries. The registered criteria use the per-query intervals and are unchanged.

\paragraph{Every-frame conditions by scene.}
\Cref{tab:supp_t1_scene} repeats \cref{tab:main} within each test scene. \Change recovers 77--83\% of the delay loss in both scenes, and its differences to the other conditions keep CIs above zero within each scene and under block resampling (\cref{tab:supp_block}).

\end{document}